\documentclass[10pt,twocolumn,letterpaper]{article}

\usepackage{cvpr}              

\usepackage{graphicx}
\usepackage{amsmath}
\usepackage{amssymb}
\usepackage{booktabs}
\usepackage{algorithm2e}
\usepackage{setspace}
\usepackage{enumitem}
\usepackage{bbm}
\usepackage{cite}
\usepackage{pifont}
\usepackage{xcolor}
\usepackage{pifont}
\usepackage{color, colortbl}
\usepackage{tabularx}
\usepackage{siunitx}
\definecolor{brightgreen}{rgb}{0.4, 1.0, 0.0}
\definecolor{LightCyan}{rgb}{0.88,1,1}
\definecolor{Gray}{gray}{0.9}
\newcommand*\colourcheck[1]{%
  \expandafter\newcommand\csname #1check\endcsname{\textcolor{#1}{\ding{52}}}%
}
\newcommand*\colourx[1]{%
  \expandafter\newcommand\csname #1x\endcsname{\textcolor{#1}{\ding{55}}}%
}
\usepackage{amsfonts}

\newcommand{\overbar}[1]{\mkern 1.5mu\overline{\mkern-1.5mu#1\mkern-1.5mu}\mkern 1.5mu}
\colourcheck{blue}
\colourcheck{green}
\colourcheck{red}
\colourcheck{brightgreen}
\colourcheck{black}
\colourcheck{gray}
\colourx{blue}
\colourx{green}
\colourx{red}
\colourx{black}
\colourx{gray}

\newcommand{\bluetext}[1]{\textcolor{blue}{#1}}

\newcommand*\yellowcircle{\includegraphics[width=.7em]{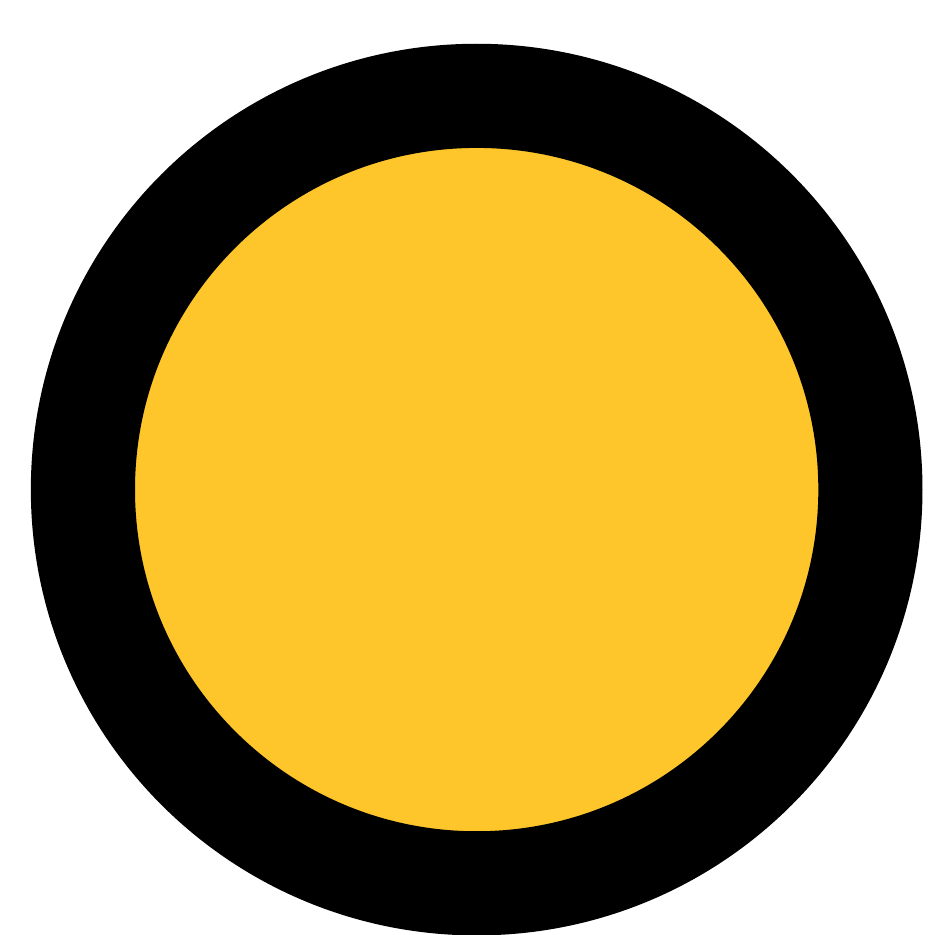}}
\newcommand*\dottedgreencircle{\includegraphics[width=.5em]{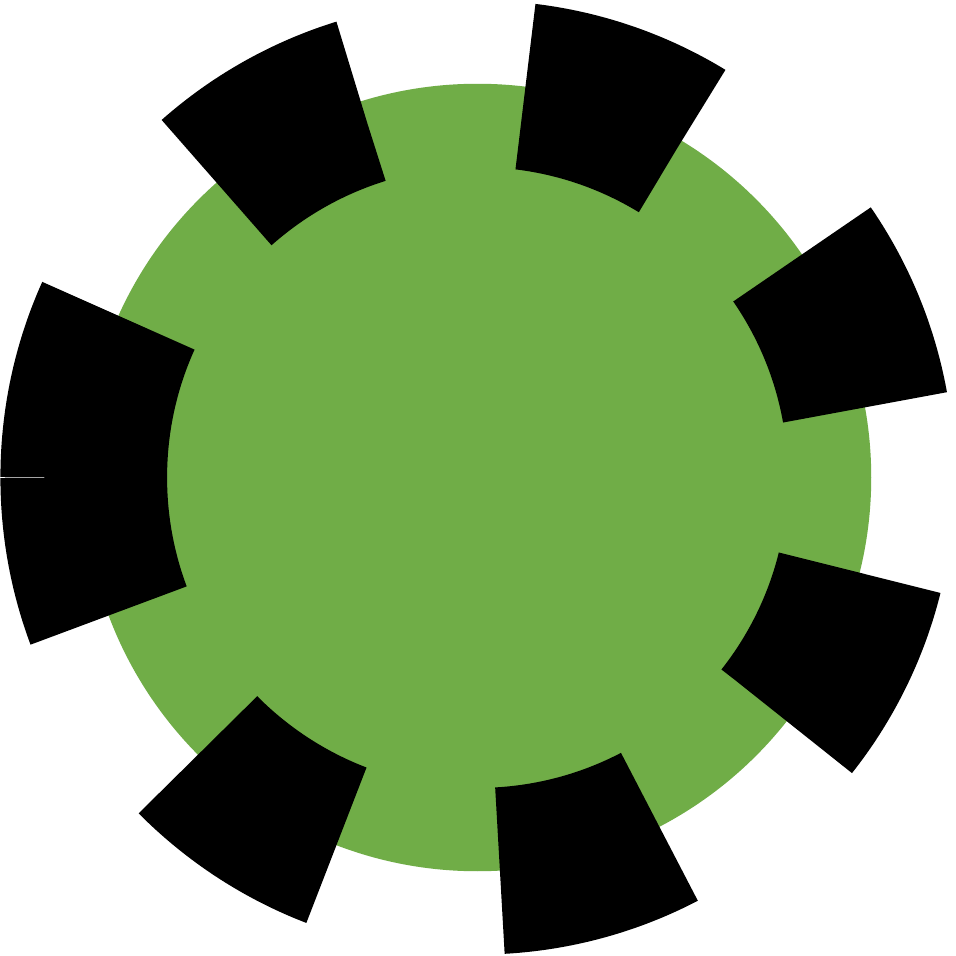}}
\newcommand*\yellowsmallcircle{\includegraphics[width=.4em]{yellow_circle.pdf}}
\newcommand*\dottedgreenbigcircle{\includegraphics[width=.7em]{dotted_green_circle.pdf}}
\usepackage[pagebackref,breaklinks,colorlinks]{hyperref}

\usepackage[capitalize]{cleveref}
\crefname{section}{Sec.}{Secs.}
\Crefname{section}{Section}{Sections}
\Crefname{table}{Table}{Tables}
\crefname{table}{Tab.}{Tabs.}

\def\cvprPaperID{827} 
\def\confName{CVPR}
\def\confYear{2023}

\begin{document}

\title{Meta-Explore: Exploratory Hierarchical Vision-and-Language Navigation\\
Using Scene Object Spectrum Grounding}

\author{
Minyoung Hwang$^1$, Jaeyeon Jeong$^1$, Minsoo Kim$^3$, Yoonseon Oh$^2$, Songhwai Oh$^1$\\
$^1$Electrical and Computer Engineering and ASRI, Seoul National University\\
$^2$Department of Electronic Engineering, Hanyang University\\
$^3$Interdisciplinary Major in Artificial Intelligence, Seoul National University\\
{\tt\footnotesize $\{$minyoung.hwang, jaeyeon.jeong$\}$@rllab.snu.ac.kr},
{\tt\footnotesize yoh21@hanyang.ac.kr},
{\tt\footnotesize $\{$goldbird5, songhwai$\}$@snu.ac.kr}\vspace{-0.45cm}
}

\maketitle

\begin{abstract}
\vspace{-0.07cm}
\fontdimen2\font=1.6pt{
  The main challenge in vision-and-language navigation (VLN) is how to understand natural-language instructions in an unseen environment. The main limitation of conventional VLN algorithms is that if an action is mistaken, the agent fails to follow the instructions or explores unnecessary regions, leading the agent to an irrecoverable path.
  To tackle this problem, we propose Meta-Explore, a hierarchical navigation method deploying an exploitation policy to correct misled recent actions. We show that an exploitation policy, which moves the agent toward a well-chosen local goal among unvisited but observable states, outperforms a method which moves the agent to a previously visited state. 
 We also highlight the demand for imagining regretful explorations with semantically meaningful clues. The key to our approach is understanding the object placements around the agent in spectral-domain.
  Specifically, we present a novel visual representation, called scene object spectrum (SOS), which performs category-wise 2D Fourier transform of detected objects. 
Combining exploitation policy and SOS features, the agent can correct its path by choosing a promising local goal. 
  We evaluate our method in three VLN benchmarks: R2R, SOON, and REVERIE. Meta-Explore outperforms other baselines and shows significant generalization performance. In addition, local goal search using the proposed spectral-domain SOS features significantly improves the success rate by 17.1\% and SPL by 20.6\% against the state-of-the-art method of the SOON benchmark. Project page: \href{https://rllab-snu.github.io/projects/Meta-Explore/doc.html}{https://rllab-snu.github.io/projects/Meta-Explore/doc.html}
  }\\
\end{abstract}\vspace{-1.0cm}

\let\thefootnote\relax\footnotetext{\scriptsize\fontdimen2\font=0.7pt This work was supported by Institute of Information \& Communications Technology Planning \& Evaluation (IITP) grant funded by the Korea government (MSIT) (No. 2019-0-01190, [SW Star Lab] Robot Learning: Efficient, Safe, and Socially-Acceptable Machine Learning). This work was partly supported by Institute of Information \& communications Technology Planning \& Evaluation (IITP) grant funded by the Korea government(MSIT) (No.2022-0-00907, Development of AI Bots Collaboration Platform and Self-organizing AI) \textit{(Corresponding authors: Yoonseon Oh and Songhwai Oh.)}}

\section{Introduction}\label{sec:intro}
\vspace{-0.1cm}
\fontdimen2\font=2.1pt{Visual navigation in indoor environments has been studied widely and shown that an agent can navigate in unexplored environments \cite{zhu2021deep}. By recognizing the visual context and constructing a map, an agent can explore the environment and solve tasks such as moving towards a goal or following a desired trajectory. With the increasing development in human language understanding, vision-and-language navigation (VLN) \cite{anderson2018vision} has enabled robots to communicate with humans using natural languages. The high degree of freedom in natural language instructions allows VLN to expand to various tasks, including (1) following fine-grained step-by-step instructions \cite{anderson2018vision, chen2019touchdown, rxr, jain-etal-2019-stay, yan2019cross, NEURIPS2021_landmark-rxr, krantz_vlnce_2020, mirowski2019streetlearn, streetnav2020, vasudevan2021talk2nav, misra-etal-2018-mapping, chi2020just} and (2) reaching a target location described by goal-oriented language instructions \cite{wu2018building, eqa, zhu2021soon, qi2020reverie, nguyen2019hanna, Nguyen_2019_CVPR, suhr-etal-2019-executing}.}
\begin{figure}[t!]{\centering\includegraphics[width=0.9\linewidth]{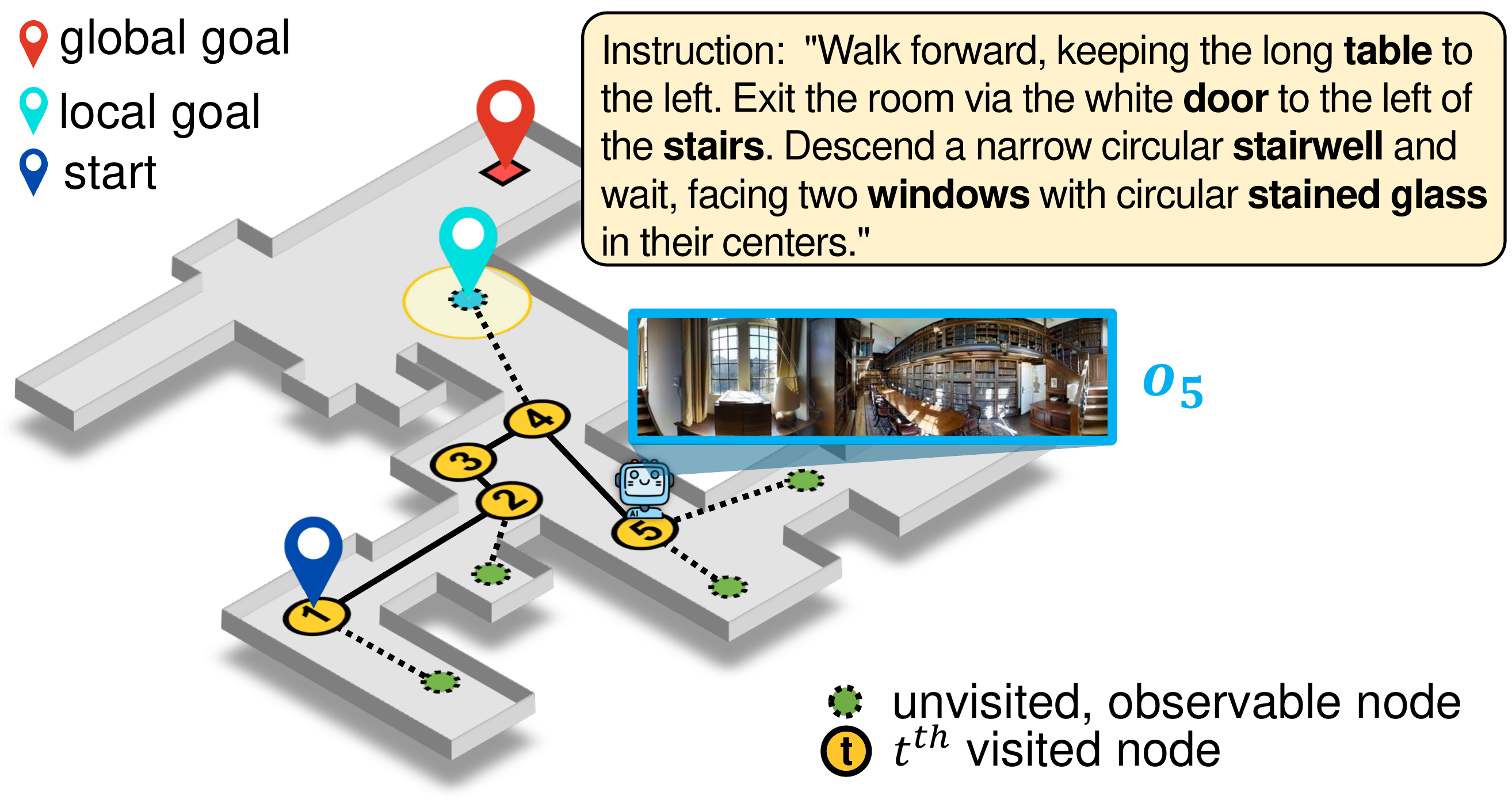}}\centering
\vspace{-0.3cm}
\caption{\protect\renewcommand{\baselinestretch}{0.9}\protect\fontdimen2\font=1.93pt\protect\small{\protect\textbf{Hierarchical Exploration.} At each episode, a natural-language instruction is given to the agent to navigate to a goal location. The agent explores the environment and constructs a topological map by recording visited nodes \yellowcircle{} and next step reachable nodes \dottedgreencircle{}. Each node consists of the position of the agent and visual features. $o_t$ denotes the observation at time $t$. The agent chooses an unvisited local goal to solve the regretful exploration problem.}}
\label{fig:overview}\vspace{-0.6cm}
\end{figure}

\fontdimen2\font=2.0pt
A challenging issue in VLN is the case when an action is mistaken with respect to the given language instruction \cite{ma2019self, ma2019regretful, zhu2020vision, chen2022think, NEURIPS2020_evolving-graph, Wang_2021_CVPR-structured-scene}. For instance, if the agent is asked to turn right at the end of the hallway but turns left, the agent may end up in irrecoverable paths.
Several existing studies solve this issue via hierarchical exploration, where the high-level planner decides when to explore and the low-level planner chooses what actions to take. If the high-level planner chooses to explore, the agent searches unexplored regions, and if it chooses to exploit, the agent executes the best action based on the previous exploration.
Prior work \cite{ma2019self, ma2019regretful, zhu2020vision} returns the agent to the last successful state and resumes exploration.
However, such methods take a heuristic approach because the agent only backtracks to a recently visited location. The agent does not take advantage of the constructed map and instead naively uses its recent trajectory for backtracking. Another recent work \cite{Wang_2021_CVPR-structured-scene} suggests graph-based exploitation, which uses a topological map to expand the action space in global planning. Still, this method assumes that the agent can directly jump to a previously visited node. Since this method can perform a jump action at every timestep, there is no trigger that explicitly decides when to explore and when to exploit. Therefore, we address the importance of time scheduling for exploration-exploitation and efficient global planning using a topological map to avoid reexploring visited regions.

\begin{figure}[t!]{\centering\includegraphics[width=1.0\linewidth]{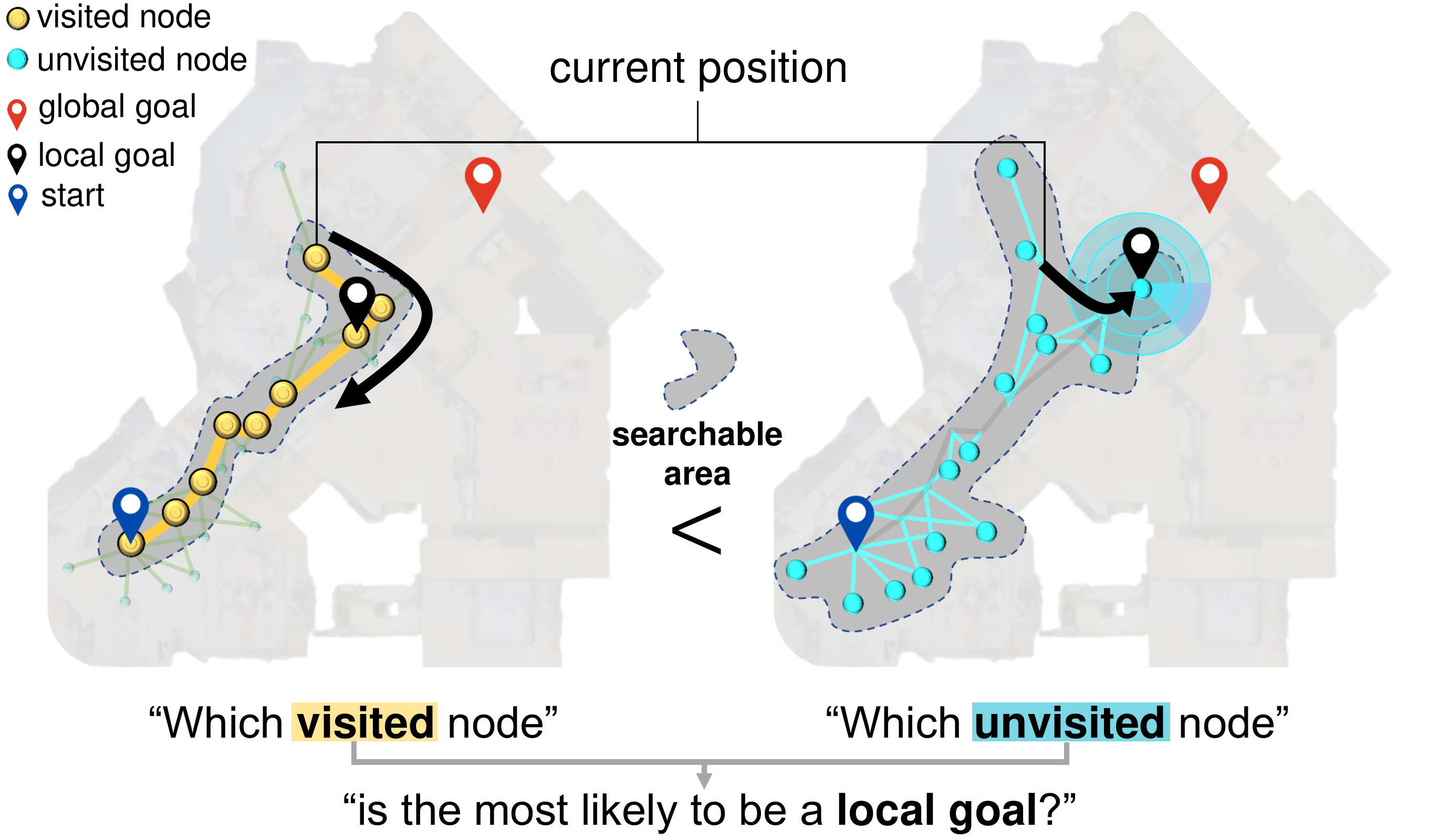}}\centering
\caption{\protect\renewcommand{\baselinestretch}{0.9}\protect\small{\protect\textbf{Local Goal Search for Exploitation.} 
The local goal is likely to be chosen as the closest node to the global goal. Existing methods only backtrack to a visited node (left). We expand the searchable area by including unvisited but reachable nodes (right).
}} \label{fig:local-goal-search}\vspace{-0.5cm}
\end{figure}

\fontdimen2\font=2.5pt{We expand the notion of hierarchical exploration by proposing Meta-Explore, which not only allows the high-level planner to choose when to correct misled local movements but also finds an unvisited state inferred to be close to the global goal. We illustrate the overview of hierarchical exploration in Figure~\ref{fig:overview}. 
Instead of backtracking, we present an exploitation method called local goal search. We show that it is more efficient to plan a path to a local goal, which is the most promising node from the unvisited but reachable nodes. We illustrate the difference between conventional backtracking and local goal search in Figure~\ref{fig:local-goal-search}. Based on our method, we show that exploration and exploitation are not independent and can complement each other: (1) to overtake regretful explorations, the agent can perform exploitation and (2) the agent can utilize the constructed topological map for local goal search.}
\fontdimen2\font=2.34pt{We also highlight the demand for imagining regretful explorations with semantically meaningful clues. Most VLN tasks require a level of understanding objects nearby the agent, but previous studies simply encode observed panoramic or object images \cite{anderson2018vision, chen2019touchdown, qi2020reverie,zhu2021soon, nguyen2019hanna, thomason2020vision, fried2018speaker, li2019robust, lu2019vilbert, guhur2021airbert, hong2021vln, pashevich2021episodic, chen2021history, ke2019tactical, ma2019self, ma2019regretful, zhu2020vision, chen2022think, NEURIPS2020_evolving-graph, Wang_2021_CVPR-structured-scene}. In this paper, we present a novel semantic representation of the scene called \textit{scene object spectrum} (SOS), which is a matrix containing the arrangements and frequencies of objects from the visual observation at each location. Using SOS features, we can sufficiently estimate the context of the environment. 
We show that the proposed spectral-domain SOS features manifest better linguistic interpretability than conventional spatial-domain visual features. 
Combining exploitation policy and SOS features, we design a navigation score that measures the alignment between a given language instruction and a corrected trajectory toward a local goal. The agent compares local goal candidates and selects a near-optimal candidate with the highest navigation score from corrected trajectories. This involves high-level reasoning related to the landmarks (e.g., bedroom and kitchen) and objects (e.g., table and window) that appear in the instructions.}
\fontdimen2\font=2.5pt

The main contributions of this paper are as follows:
\begin{itemize}

    \item We propose a hierarchical navigation method called Meta-Explore, deploying an exploitation policy to correct misled recent actions. The agent searches for an appropriate local goal instead of reversing the recent action sequence.
    \vspace{-0.2cm}
    \item In the exploitation mode, the agent uses a novel scene representation called \textit{scene object spectrum} (SOS), which contains the spectral information of the object placements in the scene. SOS features provide semantically meaningful clues to choose a near-optimal local goal and help the agent to solve the regretful exploration problem.
    \vspace{-0.2cm}
    \item We evaluate our method on three VLN benchmarks: R2R \cite{anderson2018vision}, SOON \cite{zhu2021soon}, and REVERIE \cite{qi2020reverie}. 
    The experimental results show that the proposed method, Meta-Explore, improves the success rate and SPL in test splits of R2R, SOON and val split of REVERIE. 
    The proposed method shows better generalization results compared to all baselines.
\end{itemize}

\vspace{-0.45cm}
\section{Related Work}
\vspace{-0.15cm}
\subsection{Vision-and-Language Navigation}
\vspace{-0.1cm}
In VLN, an agent encodes the natural language instructions and follows the instructions,
which can be either (1) a fine-grained step-by-step instruction the agent can follow \cite{anderson2018vision, rxr, chen2019touchdown}, (2) a description of the target object
and location \cite{qi2020reverie,zhu2021soon}, or (3) additional guidance given to the agent \cite{nguyen2019hanna, thomason2020vision}. These tasks require the agent to recognize its current location using some words in the natural-language instructions. Prior work \cite{anderson2018vision, fried2018speaker, guhur2021airbert, li2019robust, lu2019vilbert} show that an agent can align visual features to language instructions via neural networks and use the multimodal output embeddings to generate a suitable action at each timestep.
Most VLN methods utilize cross-modal attention, either with recurrent neural networks \cite{anderson2018vision, fried2018speaker}
or with transformer-based architectures \cite{li2019robust, lu2019vilbert, guhur2021airbert}. For sequential action prediction,
Hong \textit{et al.} \cite{hong2021vln} further use recurrent units inside transformer architectures, while Pashevich \textit{et al.} \cite{pashevich2021episodic} and
Chen \textit{et al.} \cite{chen2021history} use additional transformers to embed past observations and actions.
\vspace{-0.1cm}

\vspace{-1.0cm}
\subsection{Exploration-Exploitation}
\vspace{-0.15cm}
\fontdimen2\font=2.0pt
In an unseen environment, the agent must maximize the return without knowing the true value functions.
One of the solutions to this problem is to switch back and forth between exploration and exploitation \cite{march1991exploration}. In the exploration mode, the agent gathers more information about the environment. On the other hand, the agent uses information collected during exploration and chooses the best action for exploitation. Ecoffet \textit{et al.} \cite{goexplore} reduced the exploration step by archiving the states and exploring again from the successful states. 
Pislar \textit{et al.} \cite{pislar2022when} addressed the various scheduling policies and demonstrated their method on Atari games.
Recent work \cite{chaplot2020Learning, NRNS} successfully demonstrates the effectiveness of hierarchical exploration in image-goal navigation.

\begin{figure*}[ht!]{\centering\includegraphics[width=0.8\linewidth]{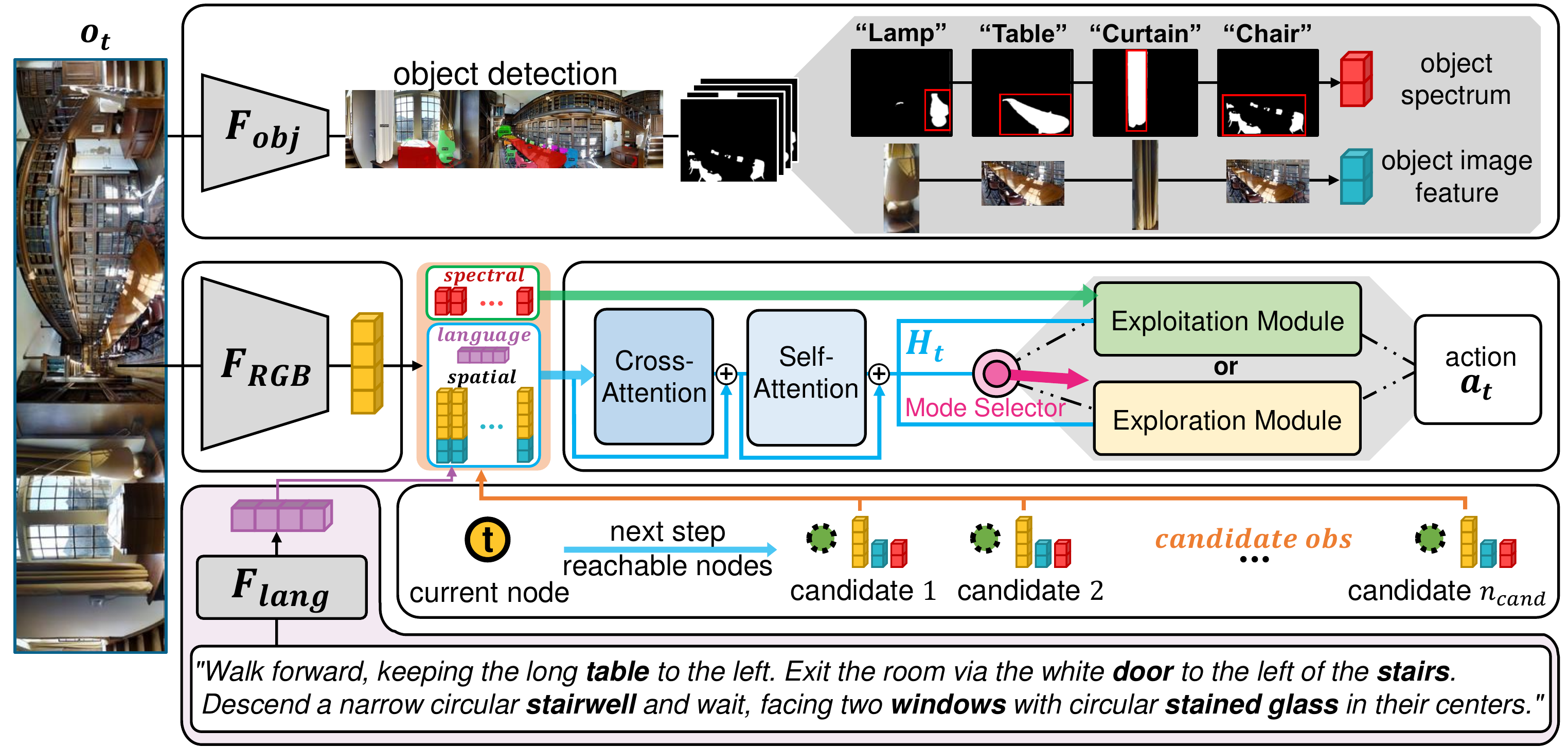}}\centering
\caption{\protect\renewcommand{\baselinestretch}{0.89}\protect\small{\protect\textbf{Network Architecture.} Three types of visual features: panoramic (yellow), object image (aquamarine), and object spectrum (red) are encoded. The color in each parenthesis denotes the color describing the corresponding feature. The cross-modal transformer encodes language and spatial visual features as hidden state $H_t$. A mode selector gives \textit{explore} or \textit{exploit} command to the agent by predicting the explore probability $P_{explore}$. The selected navigation module outputs an action $a_t$ from the possible $n_{cand}$ candidate nodes.}} \label{fig:network-architecture}\vspace{-0.5cm}
\end{figure*}

Like commonly used greedy navigation policies, VLN tasks also deal with the problem of maximizing the chance to reach the goal without knowing the ground truth map. Several VLN methods employ the concept of exploitation to tackle this problem. Ke \textit{et al.} \cite{ke2019tactical} look forward to several possible future trajectories and decide whether to backtrack or not and where to backtrack. Others \cite{ma2019self, ma2019regretful, zhu2020vision} estimate the progress to tell whether the agent becomes lost and make the agent backtrack to a previously visited location to restart exploration.
 However, previous studies do not take into account what should be done in the exploitation mode. In order to handle this problem, we propose a hierarchical navigation method which determines the scheduling between exploration and exploitation.
 \fontdimen2\font=2.5pt
\vspace{-0.2cm}

\subsection{Visual Representations}
\vspace{-0.15cm}
\fontdimen2\font=2.3pt
Popular visual encoding methods via ResNet \cite{he2016deep} and ViT \cite{dosovitskiy2020vit} can be trained to learn rotation-invariant visual features. Both methods learn to extract visual features with high information gain for global and local spatial information. The high complexity of the features leads to low interpretability of the scene and therefore requires the agent to use additional neural networks or complex processing to utilize them. On the other hand, traditional visual representation methods such as Fourier transform use spectral analysis, which is highly interpretable and computationally efficient. One drawback of the traditional methods is that they fail to maximize the information gain. Nonetheless, an appropriate use of essential information can be helpful for high-level decision making and enables more straightforward interpretation and prediction of the visual features.
One traditional navigation method, Sturz \textit{et al.} \cite{sturzl2006efficient} used Fourier transform to generate rotation-invariant visual features. However, no research has transformed the spectral information of the detected objects to represent high-level semantics from visual observations. Focusing on the fact that 2D Fourier transform can extract morphological properties of images\cite{Serra2020-mathmorphology}, we can find out the shape or structure of detected objects through 2D Fourier transform. In this paper, we decompose the object mask into binary masks by object categories and perform a 2D Fourier transform on each binary mask.
\vspace{-0.1cm}

\section{Method}
\vspace{-0.1cm}
\subsection{Problem Formulation}
\vspace{-0.1cm}
\fontdimen2\font=2.2pt
We deal with VLN in discrete environments, where the environment is given as an undirected graph $G_e=\{V, E\}$. $V$ denotes a set of $N$ navigable nodes, $\{V_i\}^N_{i=1}$, and $E$ is the adjacency matrix describing connectivity among the nodes in $V$. We denote the observation at node $V_i$ as $O_i$. The agent uses a panoramic RGB image observation $o_t$ and current node $v_t$, which are collected at time $t$. The agent either moves to a neighboring node or executes a $\texttt{stop}$ action. $a_t$ denotes the action at time $t$. The objectives of VLN are categorized as follows: (1) to follow language instructions ~\cite{anderson2018vision} and (2) to find a target object described by language instructions in a fixed time $T$ ~\cite{zhu2021soon, qi2020reverie}. We present a general hierarchical exploration method that can be applied to both tasks. We also enhance the navigation policy by extracting cross-domain visual representations from the environments, i.e., spatial-domain and spectral-domain representations. To balance the information loss and interpretability of the visual feature, we adopt multi-channel fast Fourier transform (FFT) to encode semantic masks of the detected objects into category-wise spectral-domain features.
\fontdimen2\font=2.5pt
\vspace{-0.1cm}

\subsection{Meta-Explore}
\vspace{-0.15cm}
\fontdimen2\font=2.0pt
We design a learnable hierarchical exploration method for VLN called Meta-Explore, which decides (1) when to \textit{explore} or \textit{exploit} and (2) a new imagined local goal to seek during exploitation. The overall network architecture of the proposed Meta-Explore is shown in Figure~\ref{fig:network-architecture}. Given a language instruction $L$, the agent navigates in the environment until it finds the target described in $L$. Meta-Explore consists of a mode selector and two navigation modules corresponding to two modes: exploration and exploitation. At each timestep, the mode selector chooses to explore or exploit. At $t=0$, the mode is initialized to exploration. In the \textit{exploration mode}, the agent outputs an action toward a neighboring node to move the agent toward the goal. When the mode selector recognizes that the agent is not following the instruction successfully, the mode is switched to exploitation. In the \textit{exploitation mode}, the agent seeks a new \textit{local goal} with the highest correspondence against the language instructions from the previously unvisited candidate nodes using spectral-domain visual features. The agent moves toward the local goal by planning a path. After the agent arrives at the local goal, the mode is reset to exploration. The explore-exploit switching decision occurs through the mode selector by estimating the probability to explore. The agent repeats this explore-exploit behavior until it determines that the target is found and decides to stop.
\vspace{-0.4cm}

\subsubsection{Mode Selector}
\vspace{-0.2cm}
\fontdimen2\font=2.2pt
At time $t$, the agent observes visual features about the current node $v_t$ and several reachable nodes. We call the nodes reachable at the current timestep as candidate nodes. $n_{cand}$ denotes the number of candidate nodes.
We use a cross-modal transformer with $n_L$ layers to relate visual observations to language instructions. The cross-modal transformer takes the visual features of nodes in the constructed topological map at time $t$, $G_t$, and outputs cross-modal embedding $H_t$ to encode visual observations with $L$. We concatenate location encoding and history encoding \cite{chen2022think} to the visual features as node features to consider the relative pose from $v_t$ and the last visited timestep of each node, respectively. Each word is encoded via a pretrained language encoder \cite{tan2019lxmert}, which is used for general vision-language tasks. 

The cross-modal transformer consists of cross-attention layer  \scalebox{0.9}{$\text{L2V}\_\text{Attn}(\hat{W}, \hat{V}) =\text{Softmax}({\hat{W}\Theta_q^{\dagger}(\hat{V}\Theta_k^{\dagger})^T/ \sqrt{d}})\hat{V}\Theta_v^{\dagger} \ \text{and}$} self-attention layer \scalebox{0.9}{$\text{Self}\text{Attn}(X) = \text{Softmax}$} \scalebox{0.9}{$({({X\Theta_q(X\Theta_k)^T+\!  }}$}
\scalebox{0.9}{${{A\Theta_e +b_e})/ \sqrt{d}})X\Theta_v$}, where $\hat{W}$, $\hat{V}$, $X$, and $A$ denote word, visual, node representations and adjacency matrix of $G_t$, respectively. The (query, key, value) weight matrices of self-attention and cross-attention layers are denoted as \scalebox{0.9}{$(\Theta_q, \Theta_k, \Theta_v)$} and \scalebox{0.9}{$(\Theta_q^{\dagger}, \Theta_k^{\dagger}, \Theta_v^{\dagger})$}, respectively. The final cross-modal embedding at time $t$ after passing through $n_L$ transformer layers is denoted as $H_t$.
To encourage the monotonic increasing relationship between language and visual attentions at each timestep, we define a correlation loss \scalebox{0.9}{$L_{corr} = \sum_{t=1}^T ||\text{L2V}\_\text{Attn} - I_{n_x}||_1$} for training the cross-modal transformer, where $n_x$ denotes the dimension of the $H_t$ and $I_{n_x}$ denotes an identity matrix of size $n_x \times n_x$. 

\begin{figure}[t!]{\centering\includegraphics[width=0.9\linewidth]{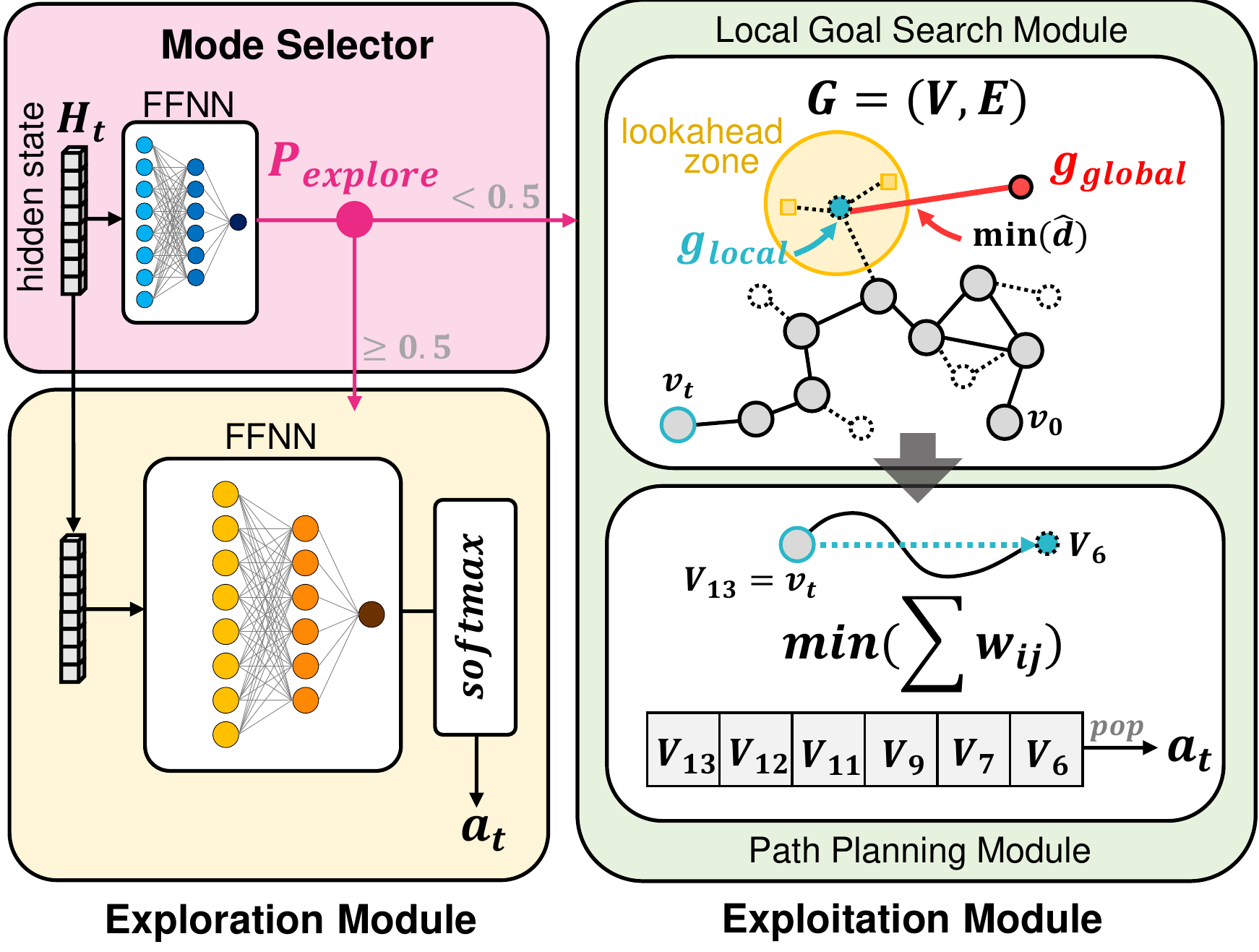}}\centering
\vspace{-0.3cm}
\caption{\protect\renewcommand{\baselinestretch}{0.9}\protect\small{\protect\textbf{Navigation Modules.} 
Mode selector estimates $P_{explore}$, i.e., the probability to explore, and chooses between exploration and exploitation modules. The selected navigation module outputs the next action $a_t$.
}}\label{fig:navigation-modules}\vspace{-0.7cm}
\end{figure}

As illustrated in Figure~\ref{fig:navigation-modules}, the mode selector estimates the probability to explore $P_{explore}$ given the cross-modal hidden state $H_t$. We denote the mode selector as $S_{mode}$ and use a two-layer feed-forward neural network. Given $H_t$, $S_{mode}$ outputs the exploration probability as $P_{explore} = 1-S_{mode}(H_t)$.
If $P_{explore}\geq0.5$, the exploration policy outputs a probability distribution for reachable nodes at the next step. At time $t+1$, the agent moves to the node with the highest probability. If $P_{explore}<0.5$, the agent determines that the current trajectory is regretful, so the agent should traverse to find a local goal, which is the most likely to be the closest node to the global goal. The exploitation policy mainly utilizes object-level features to search for the local goal with high-level reasoning. After the local goal is chosen, the path planning module outputs an action following the shortest path to the local goal.

To train the mode selector, we require additional demonstration data
other than the ground truth trajectory, such that it switches between exploration and exploitation.
We generate the demonstration data from the ground truth trajectories, with additional detours.
For the detours, we stochastically select candidate nodes other than the ground truth paths
and add the trajectory that returns to the current viewpoint.
The imitation learning loss for training the mode selector is defined as \scalebox{0.9}{$L_{mode} = \sum_{t=1}^T{\mathbbm{1}(m_t=\text{gt}_t)}$},
where $m_t$ is the mode of the agent, $0$ for exploitation and $1$ for exploration. $\text{gt}_t$ is $1$ if the current node is in the shortest ground truth trajectory and $\text{gt}_t=0$, otherwise.
\vspace{-0.3cm}

\subsubsection{Exploration Module}\label{sec:exploration-module}
\vspace{-0.2cm}
In the exploration mode, the agent follows the following sequential operations: topological map construction, self-monitoring, and an exploration policy. To improve the exploration, we adopt self-monitoring \cite{ma2019self} to predict the current progress of exploration to enhance the exploration policy itself. Prior work \cite{ma2019self, ma2019regretful} has shown that auxiliary loss using self-monitoring can regularize the exploration policy. \\
\vspace{-0.3cm}

\noindent\textbf{Topological Map Construction.}
The agent constructs graph $G_t$ by classifying nodes into two types: (1) visited nodes and (2) unvisited but observable nodes. At current time $t$, the agent at node $v_t \in \{V_i\}^N_{i=1}$ observes $N(v_t)$ neighbor nodes as next step candidates at time $t+1$. The visited nodes consist of visual features of their own and the neighboring nodes from panoramic RGB observations. The unvisited nodes can be observed only if they are connected to at least one visited node. The topological map records the positions and visual features of observed nodes at each timestep. By knowing the positions of nodes in $G_t$, the agent can plan the shortest path trajectory between two nodes.

\noindent\textbf{Self-Monitoring.}
\fontdimen2\font=1.8pt{
We use a progress monitor to estimate the current navigation progress at each episode. Self-monitoring via estimating current progress helps the agent choose the next action that can increase the progress. The estimated progress $\hat{p}_t = F_{progress}(H_t)$ is the output of a feed-forward neural network, given $H_t$ as input. We measure the ground truth progress $p_t$ as the ratio between the current distance to the goal and the shortest path length of the episode subtracted from $1$, described as $1-{\frac{d_{geo}(v_t, v_{goal})}{d_{geo}(v_0, v_{goal})}}$, where $d_{geo}(a,b)$ is the geodesic distance between $a$ and $b$. $v_0, v_t$, and $v_{goal}$ denote initial, current, and goal positions, respectively. We add progress loss \scalebox{0.9}{$L_{progress}=\sum_{t=1}^T (\hat{p_t}-p_t)^2$} to train the progress monitor while training the exploration policy.}
\fontdimen2\font=2.5pt

\fontdimen2\font=2.2pt
\noindent\textbf{Exploration Policy.}~\label{exploration-policy} The exploration policy $F_{explore}$ estimates the probability of moving to the candidate nodes at the next step. The agent chooses the action $a_t$ at time $t$ based on the estimated probability distribution among candidate nodes, described as $a_t=\arg\max_{V_i} (F_{explore}([H_t]_i))$. $F_{explore}$ is implemented via a two-layer feed-forward network with the cross-modal hidden state $H_t$ given as input. The output of $F_{explore}$ becomes a probability distribution over possible actions. To only consider unvisited nodes, we mask out the output for visited nodes. For training, we sample the next action from the probability distribution instead of choosing a node with the highest probability. We describe the training details in Section~\ref{training-details}.
\fontdimen2\font=2.5pt
\vspace{-0.4cm}

\subsubsection{Exploitation Module}\label{sec:exploitation-module}
\vspace{-0.2cm}
In the exploitation mode, the agent requires high-level reasoning with identifiable environmental clues to imagine regretful exploration cases. To find clues in an object-level manner, we present a novel visual representation by capturing object information in the spectral-domain. The novel representation is more easily predictable than spatial features such as RGB image embeddings. The agent can take advantage of the predictability by expanding the searchable area to find a local goal. We choose the local goal as the closest node to the global goal in the feature space.\\
\vspace{-0.3cm}

\begin{figure}[t!]{\centering\includegraphics[width=0.95\linewidth]{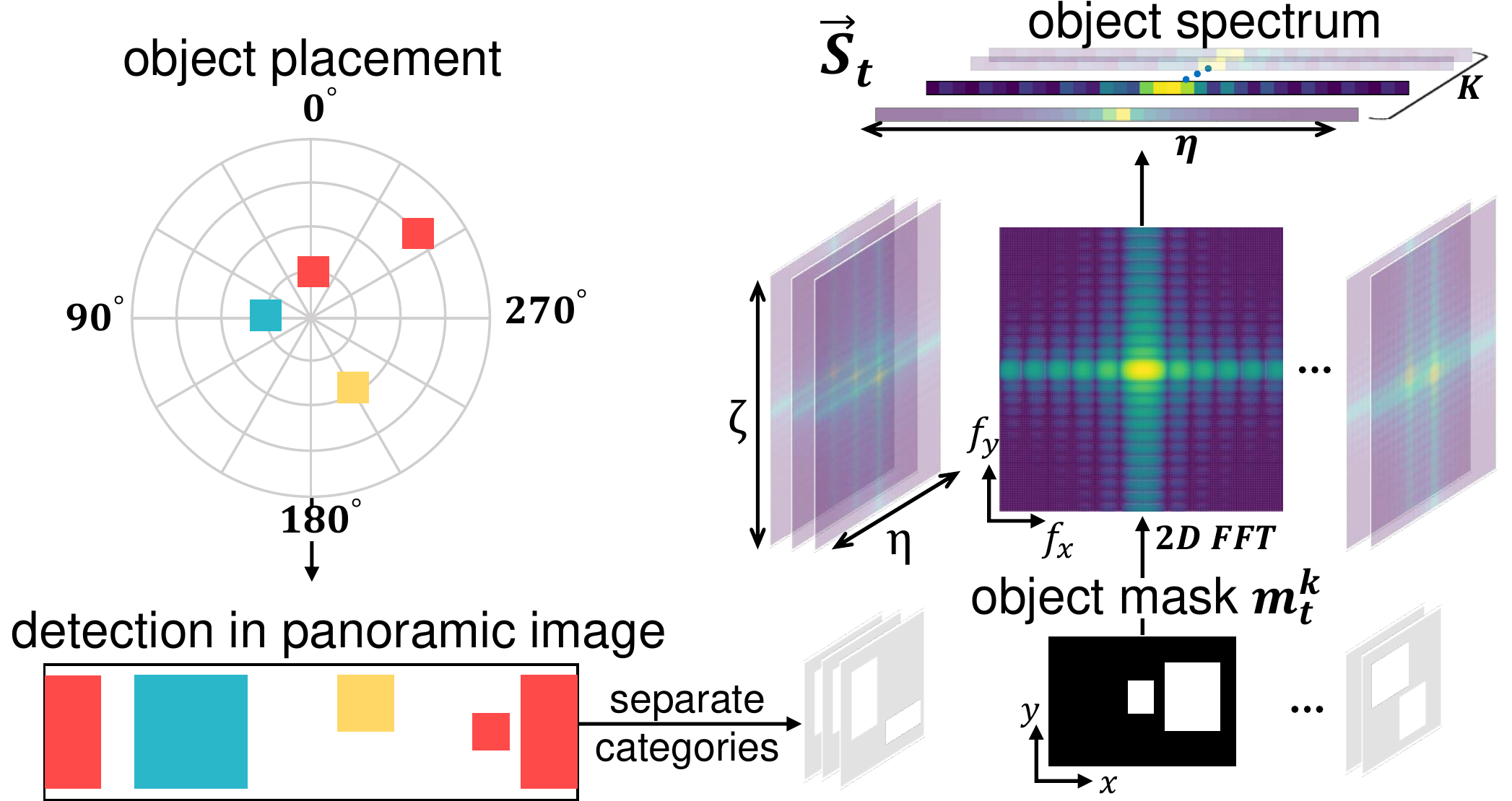}}\centering
\caption{\protect\renewcommand{\baselinestretch}{0.9}\small{{\textbf{Scene Object Spectrum (SOS).} The agent calculates \textit{scene object spectrum} (SOS) features for efficient exploitation. SOS features incorporate semantic information observed in a single panoramic image by performing category-wise 2D FFT.}\vspace{-0.4cm} }} \label{fig:scene-object-spectrum}\end{figure}

\noindent\textbf{Spectral-Domain Visual Representations}.
Common navigation policies can lead the agent toward the node with the highest similarity to the target.
However, even with a good learned policy, the agent can act in a novice manner in unseen environments.
In this paper, we seek extra information from the environment for generalizable high-level reasoning to resolve the issue. As illustrated in Figure~\ref{fig:scene-object-spectrum},
\textit{scene object spectrum} (SOS) incorporates semantic information observed in a single panoramic image by
generating a semantic mask for each object category and applying Fourier transform to each semantic mask. The semantic mask for object class $k$ at time $t$ is calculated as a binary mask $[m^k_t]_{ij}$ that detects the object at pixel $(i,\ j)$.
Suppose there are a total of $K$ object categories. When multiple objects are detected for one object category, the binary mask appears as a union of the bounding boxes of the detected objects. We define $\mathbf{FFT}$ as a channel-wise 2D fast Fourier transform
that receives $K$ binary semantic masks and outputs $K$ spectral-domain features, where $K$ is the number of object classes. 
Then, SOS feature $\vec{S_t}=[s^1_t, ..., s^K_t]^T$ can be defined as $s^k_t =\log|\mathbf{FFT}(m^k_t)|$. 
For simplicity, we perform mean pooling on the vertical spectral axis and normalize the output. The final SOS feature has shape $(K, \eta)$, where $\eta$ is the maximum horizontal frequency.

\noindent\textbf{Local Goal Search Using Semantic Clues.}\label{paragraph:local-goal-search-with-semantic-clues}
\fontdimen2\font=2.3pt
We argue that returning to a previously visited node does not guarantee the agent escapes from the local optima. Instead of backtracking to a previously visited node, the agent searches for a local goal to move towards. If the agent plans a path and moves towards the local goal, the agent does not need to repeat unnecessary actions in visited regions after the exploitation ends. Additionally, searching for a local goal takes full advantage of the topological map by utilizing the connections among the observed nodes. To expand the searchable area further, we let the agent choose the local goal from previously unvisited and unchosen candidate nodes.
\fontdimen2\font=2.5pt

To choose a local goal, we first score the corrected trajectories to measure the alignment with the language instruction $L$. We use SOS features as semantic environmental clues to estimate the navigation score $S_{nav}$ of the corrected trajectory, which is the shortest path trajectory from the initial node to the local goal in the constructed topological map.\! To simplify, we convert the language instruction into a 

\noindent list of objects $W^o =[w^o_1, ..., w^o_B]$ consisting of $B [\leq K]$ object categories (e.g., desk, cabinet, and microwave). We approximate the corresponding reference SOS features as $[\hat\delta(w^o_1), ..., \hat\delta(w^o_B)]$ where the $i^{th}$ row of $\hat\delta(w^o_k)$ is defined as \fontdimen2\font=9.0pt{$[\hat\delta(w^o_k)]_i\,=\,\mathbbm{1}(k\,=\,i)\lambda(\delta(w^o_k))\,\text{sinc}({\frac{j}{2}}\,\, -\,\, {\frac{\eta}{4}})$. $\lambda(\delta(w^o_k))$}\fontdimen2\font=2.5pt

\noindent denotes the average width of detected bounding boxes of object $w^o_k$ in the environment. A detailed approximation process is explained in the supplementary material. To simulate a corrected trajectory $\mathcal{T}'=(v'_1, ..., v'_{t'})$, we calculate the SOS features $[\vec{S}'_1, ..., \vec{S}'_{t'}]$ corresponding to the nodes in $\mathcal{T}'$. We measure the similarity between two object spectrum features via the cosine similarity of the flattened vectors. Finally, the navigation score $S_{nav}$ of $\mathcal{T}'$ is computed as:
{\footnotesize
\begin{align}\label{sos}
    \begin{split}
        S_{nav}(\mathcal{T}') = {{\sum\limits_{i=1}^{B}\sum\limits_{j=1}^{t'} ({\hat\delta(w^o_i)\over{|\hat\delta(w^o_i)|}}\cdot {\vec{S}'_j\over{|\vec{S}'_j|}})((\hat\delta(w^o_i)-\hat\delta(\overbar{w}^o))\cdot(\vec{S}'_j-\overbar{\vec{S}}'))}\over{\sqrt{{{t'}\over B}\cdot\sum\limits_{i=1}^{B} (\hat\delta(w^o_i)-\hat\delta(\overbar{w}^o))^2\sum\limits_{j=1}^{t'} (\vec{S}'_j-\overbar{\vec{S}}')^2}}},
    \end{split}
\end{align}
}\vspace{-0.4cm}

\noindent where $\hat\delta(\overbar{w}^o)$ and $\overbar{\vec{S}}'$ denote the average values of SOS features $\hat\delta({w}^o_i)$ and $\vec{S}'_j$, respectively. This equation can also be interpreted as a pseudo correlation-coefficient function between object list $W^o$ and trajectory $\mathcal{T}'$.
The exploitation policy selects the node with the highest navigation score as the local goal from the previously unvisited candidates.

\begin{figure}[t!]{\centering\includegraphics[width=1.04\linewidth]{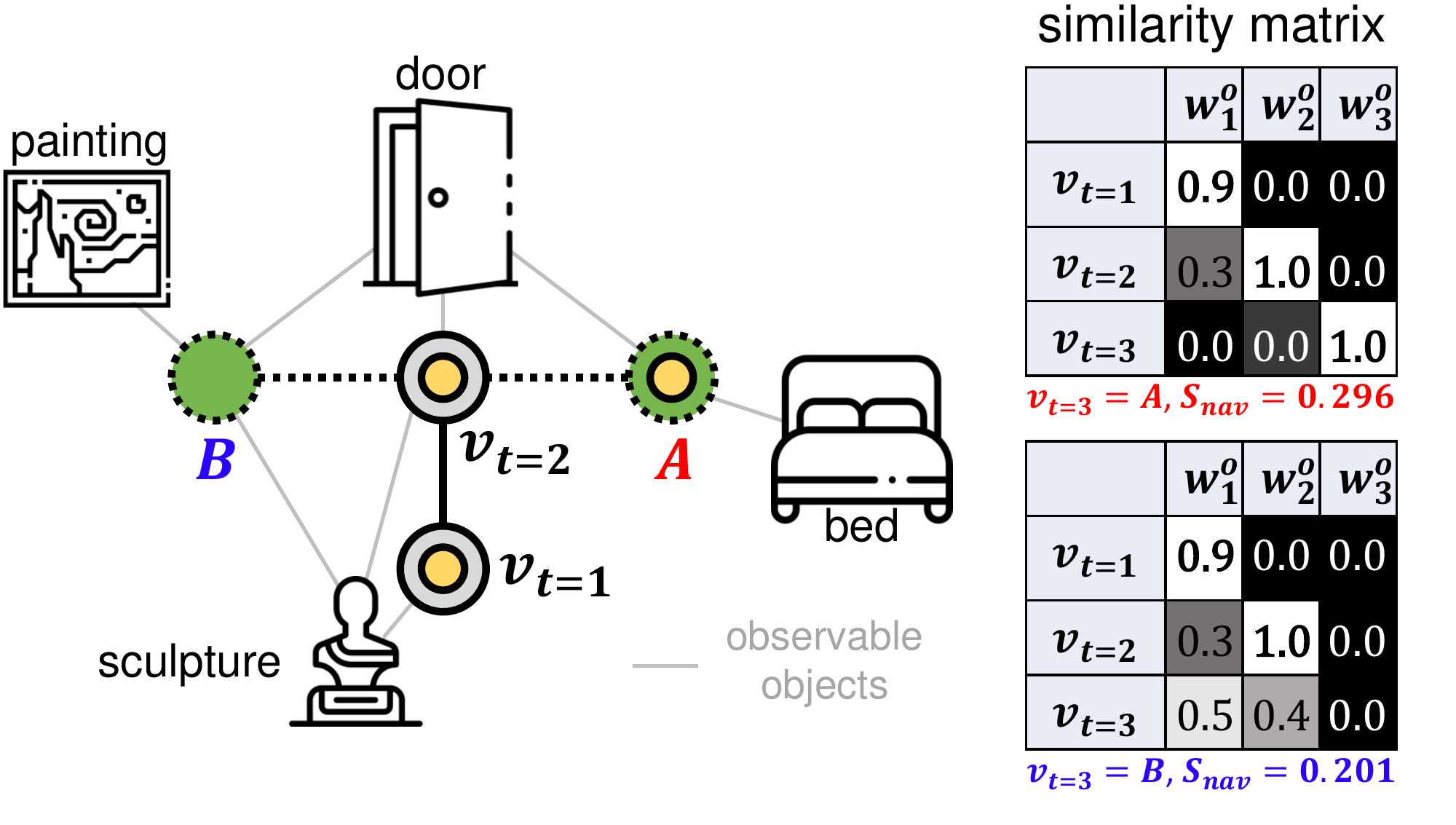}}\centering
\vspace{-0.4cm}
\caption{\protect\renewcommand{\baselinestretch}{0.89}\small{{\textbf{Toy Example.} \fontdimen2\font=1.4pt{Monotonic alignment between language instruction and visual observation is desirable. Yellow dots \yellowsmallcircle{} in the nodes describe the ground truth trajectory. Based on the node at $t=3$, the similarity matrix can show either monotonic or non-monotonic alignment between object tokens and SOS features. The green circles \dottedgreenbigcircle{} describe the possible candidates $A, B$ for next action.}} }} \label{fig:toy-example}\vspace{-0.4cm}\end{figure}

Figure~\ref{fig:toy-example} illustrates a simple scenario of entering a room. Suppose $W^o = [\text{sculpture}, \text{door}, \text{bed}]$ and the agent has to compare two trajectories $\mathcal{T}_1=(v_1, v_2, A)$ and $\mathcal{T}_1=(v_1, v_2, B)$. Each similarity matrix in Figure~\ref{fig:toy-example} has the $(t,j)$ element as the similarity between the SOS feature of $V_t$ and $\hat\delta(w^o_j)$, which is calculated as $\hat\delta(w^o_j)\cdot\vec{S}'_t$. Notably, the similarity matrix shows monotonic alignment and the navigation score is higher when the next action is chosen correctly.\vspace{-0.1cm}

\subsection{Training Details}\vspace{-0.15cm}~\label{training-details}
We use ~\cite{chen2022think} for pretraining the visual encoder with panoramic RGB observations.
We use the DAgger algorithm \cite{DBLP:dagger} to pretrain the navigation policy and the mode selector. To prevent overfitting, we iteratively perform teacher forcing and student forcing to choose the action from the exploration policy. Imitation learning loss is calculated as $L_{IL} = \sum_{t=1}^{T}-\log p(a_t^*|a_t)$ and object grounding loss is calculated as $L_{OG} = -\log p(\text{obj}^*|\text{obj}_{pred})$, where $\text{obj}^*$ denotes the ground truth and $\text{obj}_{pred}$ denotes the predicted object location. The total loss function is defined as $L_{total} = L_{mode} + L_{progress} + L_{corr} + L_{IL} + L_{OG}$.
\fontdimen2\font=1.8pt{
We further finetune the agent via A2C \cite{mnih2016asynchronous}. 
The exploration policy selects the action $a_t$ with probability $p_t^a$.
Reinforcement learning loss is defined as $L_{RL} = -\sum_t{a_t^s \log(p_t^a)A_t - \lambda\sum_t a_t^* \log(p_t^a)}$. To train the mode selector, progress monitor, and exploration policy in an end-to-end manner, we use the total loss function as $L_{fine} = L_{mode} + L_{progress} + L_{RL}$.
The exploitation policy searches the path toward the local goal from the constructed navigation graph. Thus, the exploitation policy is not learned.}
\fontdimen2\font=2.5pt

\vspace{-0.2cm}
\section{Navigation Experiments}
\label{sec:experiments}\vspace{-0.15cm}
\subsection{Experiment Settings}\vspace{-0.15cm}
We evaluate our method on three VLN benchmarks, Room-to-Room(R2R) \cite{anderson2018vision}, SOON \cite{zhu2021soon}, and REVERIE \cite{qi2020reverie}.

\noindent\textbf{R2R} evaluates the visually-grounded natural navigation performance of the agent. The agent must navigate to the predefined goal point given image observations and language instructions in an unseen environment.

\noindent\textbf{SOON} is also a goal-oriented VLN benchmark. Natural language instructions in SOON have an average length of 47 words. The agent should locate the target location and detect the location of an object to find the target object.

\noindent\textbf{REVERIE} is a goal-oriented VLN benchmark that provides natural language instruction about target locations and objects. In REVERIE, the agent is given an instruction referring to a remote object
          with an average length of 21 words. With this instruction and a panoramic observation from the environment, the agent should navigate to the location the
          instruction describes and find the correct object bounding box among the predefined object bounding boxes.
\vspace{-0.5cm}
\subsection{Evaluation Metrics}
\vspace{-0.15cm}
\subsubsection{Navigation performance}
\vspace{-0.15cm}
We evaluate algorithms using the trajectory length (TL), success rate (SR),
and success weighted by inverse path length (SPL)
\cite{anderson2018evaluation}, and oracle success rate (OSR) for the navigation performance comparison. An episode is recorded as a success if the agent takes a $\texttt{stop}$ action within 3 m of the target location. TL is the average path length in meters. SR is denoted as the number of successes divided by the total number of episodes, $M$. SPL is calculated as ${1\over M}\sum_{i=1}^M S_i {l_i\over{\max(p_i, l_i)}}$, where $S_i$ denotes the success as a binary value. $p_i$ and $l_i$ denote the shortest path and actual path lengths for the $i^{th}$ episode. OSR uses the oracle stop policy instead of the stop policy of the agent.\vspace{-0.1cm}
\begin{table*}[ht]
\renewcommand{\arraystretch}{0.92}
\setlength{\tabcolsep}{4.95pt}
\fontsize{8}{9}\selectfont
\begin{tabular}{c|c|c|cccc|cccc|cccc}
\toprule
\textbf{Methods} & \textbf{Memory} & \textbf{Exploit} &
\multicolumn{4}{c|}{\textbf{Val Seen}} & \multicolumn{4}{c|}{\textbf{Val Unseen}} & \multicolumn{4}{c}{\textbf{Test Unseen}} \\
 &  &  & SR$\uparrow$ & SPL$\uparrow$ & TL$\downarrow$ & NE$\downarrow$ & SR$\uparrow$ & SPL$\uparrow$ & TL$\downarrow$ & NE$\downarrow$ & SR$\uparrow$ & SPL$\uparrow$ & TL$\downarrow$ & NE$\downarrow$ \\ \hline\hline
Random & - & - & 16 & - & 9.58 & 9.45 & 16 & - & 9.77 & 9.23 & 13 & 12 & 9.89 & 9.79 \\
Human & - & - & - & - & - & - & - & - & - & - & 11.85 & 1.61 & 86 & 76 \\ \hline
Seq2Seq \cite{anderson2018vision} & Rec & \redx & 6.0 & 39 & 11.33 & - & 22 & - & \textbf{8.39} & 7.84 & 20 & 18 & \textbf{8.13} & 7.85 \\
VLN$\circlearrowright$BERT \cite{hong2021vln} & Rec & \redx & 72 & 68 & \textbf{11.13} & 2.90 & 63 & 57 & 12.01 & 3.93 & 63 & 57 & 12.35 & 4.09 \\
\rowcolor{Gray}SMNA$^\dagger$ \cite{ma2019self} & Rec & \textbf{\bluetext{homing}} & 69 & 63 & 11.69 & 3.31 & 47 & 41 & 12.61 & 5.48 & 61 & 56 & - & 4.48 \\
\rowcolor{Gray}Regretful-Agent \cite{ma2019regretful} & Rec & \textbf{\bluetext{homing}} & 69 & 63 & - & 3.23 & 50 & 41 & - & 5.32 & 48 & 40 & - & 5.69 \\
\rowcolor{Gray}FAST (short) \cite{ke2019tactical} & Rec & \textbf{\bluetext{homing}} & - & - & - & - & 56 & 43 & 21.17 & 4.97 & 54 & 41 & 22.08 & 5.14 \\
\rowcolor{Gray}FAST (long) \cite{ke2019tactical} & Rec & \textbf{\bluetext{homing}} & 70 & 04 & 188.06 & 3.13 & 63 & 02 & 224.42 & 4.03 & 61 & 03 & 196.53 & 4.29 \\
HAMT-e2e \cite{chen2021history} & Seq & \redx & 76 & 72 & 11.15 & 2.51 &  66 & 61 & 11.46 & \textbf{2.29} & 65 & 60 & 12.27 & 3.93 \\
DUET \cite{chen2022think} & Top. Map & \redx & 79 & 73 & 12.32 & 2.28 & \textbf{72} & 60 & 13.94 & 3.31 & 69 & 59 & 14.73 & 3.65 \\
\rowcolor{Gray}SSM \cite{Wang_2021_CVPR-structured-scene} & Top. Map & \textbf{\bluetext{jump}} & 71 & 62 & 14.7 & 3.10 & 62 & 45 & 20.7 & 4.32 & 61 & 46 & 20.4 & 4.57 \\ \hline
\rowcolor{LightCyan}\textbf{Meta-Explore (Ours)} & Top. Map & \textbf{\bluetext{local goal}} & \textbf{81} & \textbf{75} & 11.95 & \textbf{2.11} & \textbf{72} & \textbf{62} & 13.09 & 3.22 & \textbf{71} & \textbf{61} & 14.25 & \textbf{3.57} \\
\bottomrule
\end{tabular}
\vspace{-0.2cm}
\caption{\small Comparison and evaluation results of the baselines and our model in the R2R Navigation Task. 
}
\vspace{-0.4cm}
\begin{center}{\footnotesize Gray shaded rows describe hierarchical navigation baselines. Three memory types: Rec\! (recurrent), Seq\! (sequential), and Top. Map\! (topological map)}\end{center}
\label{tab:r2r-baseline_results}\vspace{-0.5cm}
\end{table*}
\begin{table*}[ht]
\renewcommand{\arraystretch}{0.92}
\setlength{\tabcolsep}{4.35pt}
\fontsize{8}{9}\selectfont
\begin{tabular}{c|c|c|ccc|c|ccc|c|ccc|c}
\toprule
\textbf{Methods} & \textbf{Memory} & \textbf{Exploit} &
\multicolumn{4}{c|}{\textbf{Val Seen Instruction}} & \multicolumn{4}{c|}{\textbf{Val Seen House}} & \multicolumn{4}{c}{\textbf{Test Unseen House}} \\
 &  & & SR$\uparrow$ & SPL$\uparrow$ & OSR$\uparrow$ & FSPL$\uparrow$ & SR$\uparrow$ & SPL$\uparrow$ & OSR$\uparrow$ & FSPL$\uparrow$ & SR$\uparrow$ & SPL$\uparrow$ & OSR$\uparrow$ & FSPL$\uparrow$ \\ \hline\hline
Human & - & - & - & - & - & - & - & - & - & -  & 90.4 & 59.2 & 91.4 & 51.1 \\ \hline
Random & Rec & \redx & 0.0 & 1.5 & 0.1 & 1.4 & 0.1 & 0.0 & 0.4 & 0.9 & 2.1 & 0.4 & 2.7 & 0.0 \\
Speaker-Follower \cite{fried2018speaker} & Rec & \redx & 97.9 & 97.7 & 97.8 & 24.5 & 61.2 & 60.4 & 69.4 & \textbf{9.1}  & 7.0 & 6.1 & 9.8 & 0.6 \\
RCM \cite{wang2019reinforced} & Rec & \redx & 84.0 & 82.6 & 89.1 & 10.9 & 62.4 & 60.9 & 72.7 & 7.8 & 7.4 & 6.2 & 12.4 & 0.7 \\
AuxRN \cite{zhu2020vision} & Rec & \redx & 98.4 & 97.4 & \textbf{98.7} & 13.7 & 68.8 & \textbf{67.3} & \textbf{78.5} & 8.3  & 8.1 & 6.7 & 11.0 & 0.5 \\
GBE w/o GE & Top. Map & \redx & 89.5 & 88.3 & 91.8 & 24.2 & 62.5 & 60.8 & 73.0 & 6.7 & 11.4 & 8.7 & 18.8 & 0.8 \\
GBE \cite{zhu2021soon} & Top. Map & \redx & 98.4 & 97.9 & 98.6 & \textbf{44.2} & \textbf{76.3} & 62.5 & 64.1 & 7.3  & 11.9 & 10.2 & 19.5 & 1.4 \\
GBE$^\dagger$ & Top. Map & \redx & - & - & - & - & 19.5 & 13.3 & 28.5 & 1.2  & 12.9 & 9.2 & 21.5 & 0.5 \\
DUET \cite{chen2022think} & Top. Map & \redx & 94.0 & 91.6 & 90.0 & 31.1 & 36.3 & 22.6 & 50.9 & 3.8  & 33.4 & 21.4 & 43.0 & \textbf{4.2} \\ \hline
\rowcolor{LightCyan}\textbf{Meta-Explore (Ours)} & Top. Map & \textbf{\bluetext{local goal}} & \textbf{100.0} & \textbf{99.1} & 96.0 & 33.9 & 44.7 & 34.8 & 52.7 & 8.9 & \textbf{39.1} & \textbf{25.8} & \textbf{48.7} & 4.0 \\
\bottomrule
\end{tabular}
\vspace{-0.2cm}
\caption{\small Comparison and evaluation results of the baselines and our model in the SOON Navigation Task.
}
\label{tab:soon-baseline_results}\vspace{-0.5cm}
\end{table*}
\vspace{-0.3cm}
\subsubsection{Object grounding performance}
\vspace{-0.15cm}
\fontdimen2\font=2.0pt{We also evaluate the object grounding performance of the agent by the success rate of finding the target object (FSR) and the target finding success weighted by inverse path
length (FSPL)\footnote{Identical with its original term, Remote Grounding Success (RGS).} \cite{zhu2021soon, qi2020reverie}. FSPL is calculated as
    $\text{FSPL} = \frac{1}{N}\sum_{i=1}^N{S_i^{nav}S_i^{loc}\cdot {l_i^{nav}}/{\max(l_i^{nav}, l_i^{gt})}}$, where $S_i^{nav}$
    is whether the agent navigates to the target, $S_i^{loc}$ is whether the agent finds a target object bounding box, and
    $l_i^{nav}$ and $l_i^{gt}$ are the navigation trajectory length and ground truth trajectory length, respectively.
}
\fontdimen2\font=2.5pt
\vspace{-0.1cm}

\subsection{Baselines and Implementation Details}
\vspace{-0.15cm}
We compare our method with several other baselines as follows. For each task, we compare our method with a number of baselines that use various types of memory (recurrent, sequential, and topological map). For methods implemented with a hierarchical navigation framework, we compare the specific exploitation methods: homing, jump, and local goal search. Homing makes the agent backtrack, and jump makes the agent jump to a previously visited node.
The hyperparameters and detailed model architecture of Meta-Explore are described in the supplementary material.
\fontdimen2\font=2.5pt
\vspace{-0.15cm}
\subsection{Comparison with Navigation Baselines}
\vspace{-0.15cm}
We compare our method with navigation baselines\footnote{$^\dagger$ indicates reproduced results.}. We focus on the success rate and SPL. Rendered results and detailed analyses with other evaluation metrics are provided in the supplementary material.

\noindent\textbf{R2R.}
Table~\ref{tab:r2r-baseline_results} compares the proposed Meta-Explore with baselines for the R2R navigation task. We categorize the baseline methods based on the type of constructed memory and the type of exploitation. Our method outperforms other exploration-only baselines over all types of validation and test splits in success rate and SPL. Compared with hierarchical baselines SMNA \cite{ma2019self}, Regretful-Agent \cite{ma2019regretful}, FAST \cite{ke2019tactical}, and SSM \cite{Wang_2021_CVPR-structured-scene}, Meta-Explore improves success rate and SPL by at least 16.4\% and 8.9\%, respectively. The main difference is that Meta-Explore constructs a topological map during exploration and uses the map for local goal search in exploitation. On the contrary, homing exploitation policies in SMNA, Regretful-Agent, and FAST only rely on the current trajectory, instead of taking advantage of the constructed memory. Jump exploitation in SSM uses a topological map to search a successful previous node, but it makes an unrealistic assumption that the agent can directly jump to a previously visited distant node and unfairly saves time. In our approach, we plan a path to the local goal based on the topological map. The experiment results reveal that even if we design a hierarchical navigation framework, exploration and exploitation are not entirely separate but they can complement each other.

\noindent\textbf{SOON, REVERIE.}
Table~\ref{tab:soon-baseline_results} compares Meta-Explore with baselines in the SOON navigation task. While the proposed method does not improve performance in val seen split, Meta-Explore outperforms other baselines in the test unseen split of SOON for success rate by 17.1\% and SPL by 20.6\%. The result implies that for the goal-oriented VLN task, high performance in train or val seen splits can be the overfitted result. Because the agent can be easily overfitted to the training data, making a generalizable model or providing a deterministic error-correction module for inference is essential. Meta-Explore chooses the latter approach by correcting the trajectory via exploitation in regretful cases. The evaluation results in the REVERIE navigation task are described in the supplementary material. Meta-Explore shows improvement in the val split of REVERIE for success rate and SPL, but the improvement in the test split is lower than the results in R2R and SOON. We found 252 meaningless object categories (e.g., verbs, adjectives, and prepositions) and 418 replaceable object categories (e.g., typographical errors and synonyms) in the REVERIE\footnote{10.7\% and 41.2\% of a total of 46,476 words in the bounding box dataset correspond to meaningless and replaceable object categories, respectively.} dataset. Because our exploitation method utilizes object-based parsing of the given instruction to match with the detected object categories, the effectiveness of the proposed method is lessened due to inaccuracies and inconsistencies in the dataset. We expect to have higher performance if the mistakes in the dataset are fixed.
\vspace{-0.2cm}
\subsection{Local Goal Search using SOS Features}
\vspace{-0.15cm}
To discuss the significance of modeling exploitation policy, we conduct specific experiments about choosing the local goal for R2R and SOON. We evaluate our method using different types of local goal search, as shown in Table~\ref{tab:localgoal_compare_results-r2r} and~\ref{tab:localgoal_compare_results-soon}. Oracle denotes a method which selects a local goal using the ground truth trajectory. The performance of the oracle provides the achievable performance for each dataset. The results imply that local goal search using either spatial or spectral visual representations is more effective than random local goal search. The results show that local goal search using spectral visual representations, i.e., SOS features, lead the agent to desirable nodes the most. We also compare local goal search with homing and the difference between the performance of the two methods is most noticeable in the test split of the SOON navigation task. As shown in Table~\ref{tab:localgoal_compare_results-soon}, choosing the local goal with only spatial-domain features, the navigation performance does not improve compared to homing.
On the contrary, spectral-domain local goal search shows significant improvement against homing by 10.4\% in success rate, 34.5\% on SPL, and 27.4\% on FSPL. The results imply that using spectral-domain SOS features helps high-level decision making, thereby enhancing the navigation performance. To further show the effectiveness of SOS features, we provide sample local goal search scenarios in the supplementary material.
\vspace{-0.2cm}
\begin{table}[htb!]
\renewcommand{\arraystretch}{0.9}
\setlength{\tabcolsep}{2.5pt}
\fontsize{7}{8}\selectfont
\begin{tabular}{c|ccccc|ccccc}
\toprule
\textbf{Local} & \multicolumn{5}{c|}{\textbf{Val Seen}} & \multicolumn{5}{c}{\textbf{Val Unseen}}\\
\textbf{Goal} & SR$\uparrow$ & SPL$\uparrow$ & OSR$\uparrow$ & TL$\downarrow$ & NE$\downarrow$ & SR$\uparrow$ & SPL$\uparrow$ & OSR$\uparrow$ & TL$\downarrow$ & NE$\downarrow$ \\ \hline\hline
Oracle & 81.88 & 74.12 & 87.46 & 13.06 & 1.93 & 75.95 & 62.53 & 84.16 & 14.00 & 2.71 \\ \hline
Random  & 79.33 & 72.67 & 85.31 & 13.19 & 2.22 & 70.97 & 59.45 & 80.16 & 14.92 & 3.34 \\
Homing  & 80.22 & 73.63 & 85.60 & 12.51 & 2.14 & 71.65 & 60.60 & 80.33 & 13.91 & 3.26 \\
Spatial  & 79.63 & 73.14 & 85.60 & 12.99 & 2.22 & 71.56 & 60.01 & 80.33 & 14.90 & 3.27 \\
\rowcolor{LightCyan}Spectral  & \textbf{80.61} & \textbf{75.15} & \textbf{85.80} & \textbf{11.95} & \textbf{2.11} & \textbf{71.78} & \textbf{61.68} & \textbf{80.76} & \textbf{13.09} & \textbf{3.22} \\
\bottomrule
\end{tabular}\vspace{-0.2cm}
\caption{\small Comparison of Exploitation Policies. (R2R)
}
\label{tab:localgoal_compare_results-r2r}\vspace{-0.6cm}
\end{table}
\begin{table}[htb!]
\renewcommand{\arraystretch}{0.9}
\setlength{\tabcolsep}{4.06pt}
\fontsize{7}{8}\selectfont
\begin{tabular}{c|cccc|cccc}
\toprule
\textbf{Local} & \multicolumn{4}{c|}{\textbf{Val Seen House}} & \multicolumn{4}{c}{\textbf{Test Unseen House}}\\
\textbf{Goal} & SR$\uparrow$ & SPL$\uparrow$ & OSR$\uparrow$ & FSPL$\uparrow$ & SR$\uparrow$ & SPL$\uparrow$ & OSR$\uparrow$ & FSPL$\uparrow$ \\ \hline\hline
Oracle  & 54.42 & 37.96 & 63.72 & 11.01 & 48.38 & 28.45 & 62.98 & 4.74 \\ \hline
Random   & 24.78 & 11.97 & 34.96 & 3.08 & 24.19 & 7.41 & 35.84 & 1.29 \\
Homing   & 42.04 & 27.72 & 48.23 & \textbf{10.18} & 35.40 & 19.18 & \textbf{51.62} & 3.14 \\
Spatial  & 32.30 & 11.60 & 39.38 & 1.90 & 26.11 & 10.58 & 39.23 & 1.43 \\
\rowcolor{LightCyan}Spectral & \textbf{44.69} & \textbf{34.84} & \textbf{52.65} & 8.89 & \textbf{39.09} & \textbf{25.80} & 48.67 & \textbf{4.01} \\
\bottomrule
\end{tabular}\vspace{-0.2cm}
\caption{\small Comparison of Exploitation Policies. (SOON)
}
\label{tab:localgoal_compare_results-soon}\vspace{-0.8cm}
\end{table}
\subsection{Ablation Study}
\vspace{-0.2cm}
We conduct an ablation study to compare the proposed method against language-triggered hierarchical exploration. Results in the supplementary material show that among the three representation domains, spatial, spectral, and language, the spectral-domain features enhance navigation performance the most. Additionally, to implicate further applications of Meta-Explore in continuous environments, we evaluate our method on the photo-realistic Habitat \cite{szot2021habitat} simulator to solve image-goal navigation and vision-and-language navigation tasks. Implementation details and results are included in the supplementary material. Results show that our method outperforms baselines in both tasks.
\vspace{-0.2cm}

\section{Conclusion}
\vspace{-0.2cm}
We have proposed Meta-Explore, a hierarchical navigation method for VLN, by correcting mistaken short-term actions via efficient exploitation. In the exploitation mode, the agent is directed to a local goal which is inferred to be the closest to the target. A topological map constructed during exploration helps the agent to search and plan the shortest path toward the local goal. To further search beyond the frontier of the map, we present a novel visual representation called \textit{scene object spectrum} (SOS), which compactly encodes the arrangements and frequencies of nearby objects. Meta-Explore achieves the highest generalization performance for test splits of R2R, SOON, and val split of REVERIE navigation tasks by showing less overfitting and high success rates. We plan to apply Meta-Explore for VLN tasks in continuous environments in our future work.

{\small
\bibliographystyle{unsrt}
\bibliography{ref.bib, egbib.bib}
}
\clearpage

\appendix
\twocolumn[\section*{\centering Supplementary Material for ``Meta-Explore: Exploratory Hierarchical\\ Vision-and-Language Navigation Using Scene Object Spectrum Grounding"}]

We provide additional details and analyses of the proposed method in this supplementary material. Section~\ref{sec:model-details} provides model details. Section~\ref{sec:experiment-setup} provides detailed settings and data preprocessing for experiments. Section~\ref{sec:navigation-experiments} provides evaluation results with detailed analyses. Section~\ref{sec:ablation-study} provides implementation details and detailed results for the ablation study.

\section{Model Details}\label{sec:model-details}
\subsection{Algorithm Details}
Algorithm~\ref{alg:meta-explore} summarizes the overall hierarchical exploration process. 
 The mode selector supervises the process and chooses whether the agent should explore or exploit at each time step.

\subsection{Exploitation Module}\label{sec:suppl-exploitation-module}
\subsubsection{Reference SOS Features.}
\begin{figure}[t!]{\centering\includegraphics[width=0.9\linewidth]{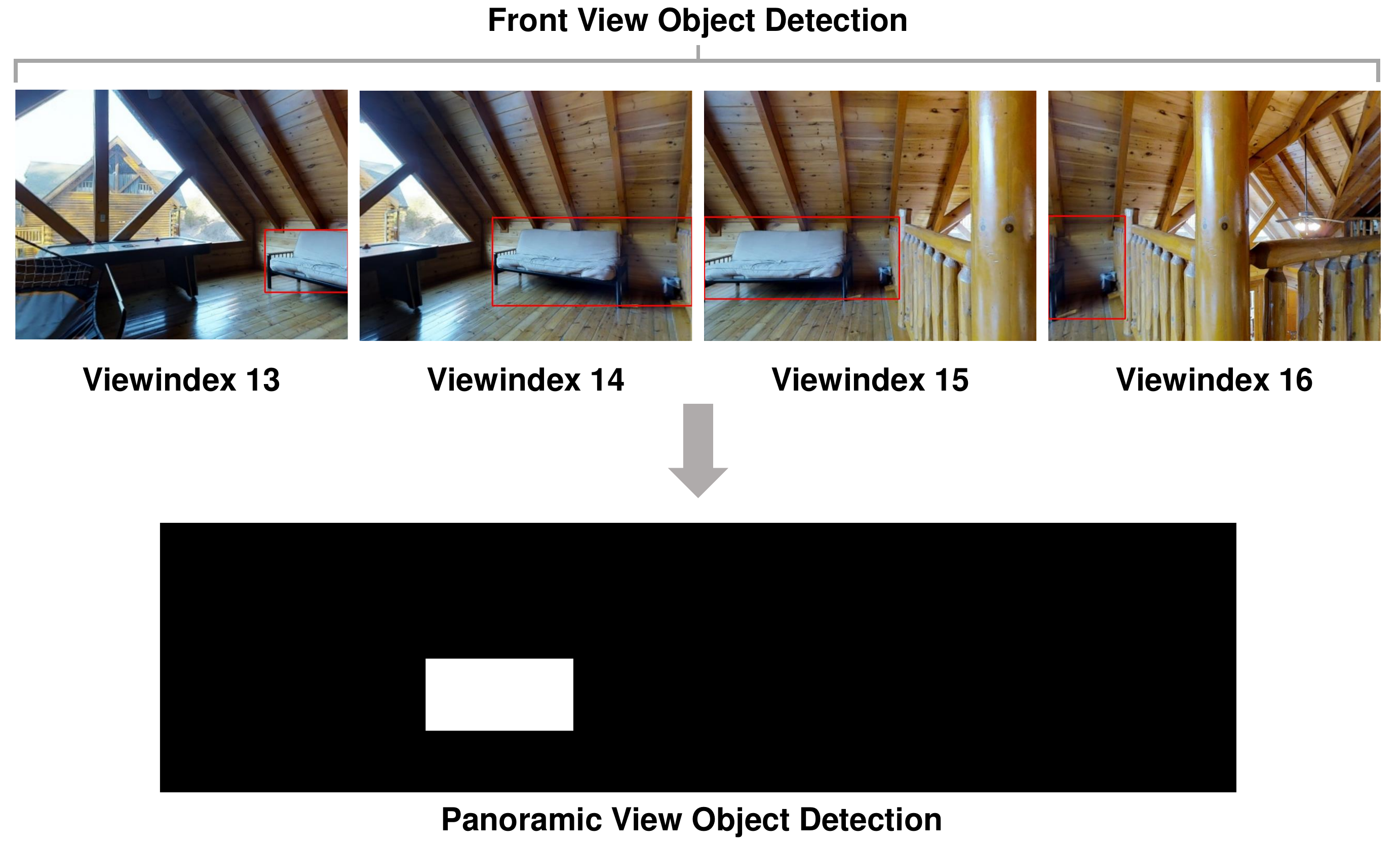}}\centering
\caption{{\textbf{Bounding box coordinate transformation.} Front-view visual observations from different angles at the same location. Each bounding box shows the `chair' detection. We use coordinate transformation to convert coordinates into panoramic view.
}}\label{fig:bbox-coordinate-transformation}
\end{figure}

\begin{figure}[t!]{\centering\includegraphics[width=1.0\linewidth]{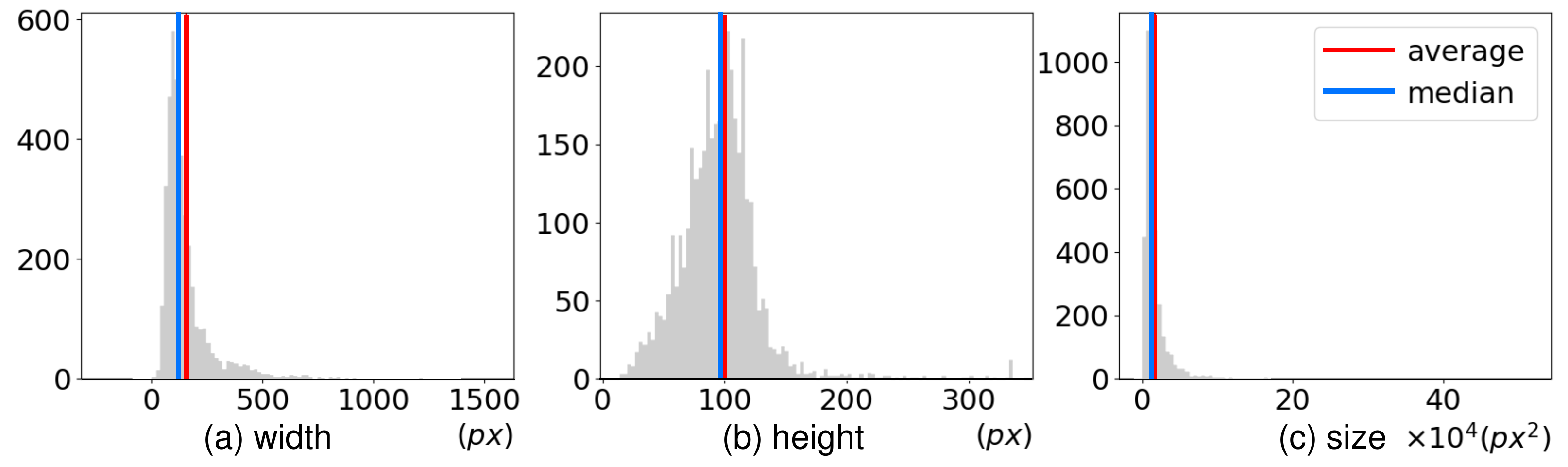}}\centering
\caption{{\textbf{Bounding box statistics.} We collect width, height, and size of detected bounding boxes. The histograms show statistics for `chair' objects. Yellow line and red line show the median and average values of each distribution, respectively.
}}\label{fig:bbox-size} \vspace{-0.2cm}
\end{figure}

    In the proposed method, we approximate the reference SOS feature of an object token by using prior information about objects in the training data. For instance, for the `chair' object, we collected the widths and heights of the detected bounding boxes as shown in Figure~\ref{fig:bbox-coordinate-transformation}. Figure~\ref{fig:bbox-size} shows two representative values: median and mean for each distribution. We choose the median values, which minimizes the L1 error, to represent the reference bounding box of each object.
To generate rotation-invariant SOS features, we convert the four vertices of the bounding box detected from the front view image of size $640\times 480$ to the vertices of a bounding box detected from the panoramic view image of size $2048\times512$ using coordinate transformations. To simplify the implementation, we assume that the converted bounding box has a rectangle shape with the vertices transformed into coordinates in a panoramic view. The reference SOS feature is calculated as the logarithmic magnitude of the Fourier transform of the panoramic mask with mean pooling on the vertical spectral axis. Considering that the shift in the spatial-domain only affects the phase of the Fourier transform, the location of a reference bounding box does not matter.

\RestyleAlgo{ruled}
\begin{algorithm}[htb]
\SetAlgoLined
\DontPrintSemicolon
        \caption{Meta-Explore} \label{alg:meta-explore}
        ${{P}}_{explore} \gets 1$\;
        $Success \gets False$\;
        Initialize $G_t$ and node features\;
        \While{$t < T$}{
        Update $G_t$\;
        Update node features\;
        $H_t \gets$ cross-modal embedding at time $t$\;
    \eIf{$P_{explore}\geq0.5$}{
        $a_t \gets \arg\max_{V_i} (F_{explore}([H_t]_i))$\;
        $\hat{p}_t \gets F_{progress}(H_t)$\;
        $t \gets t+1$\;
    }{  
        $V_{local} \gets $ unvisited but observed nodes in $G_t$\; 
        $v_{local} \gets {\arg\max}_{v' \in V_{local}}{(S_{nav}(\tau'(v_0, v')))}$\;
        $\tau \gets Path Planning(v_t, v_{local})$\;
        \While{not arrived at $v_{local}$}{$a_t \gets pop(\tau)$\;$t \gets t+1$\;}
    }
    $P_{explore} \gets 1-S_{mode}(H_t)$\;
    \lIf{$a_t$ is \texttt{stop} and $d(v_t, v_{goal})<d_{success}$}{$Success \gets True$}}
\end{algorithm}

\begin{table*}[]
\centering
\setlength{\tabcolsep}{3.4pt}
\fontsize{7}{10}\selectfont
\begin{tabular}{c|l|l|c}
\toprule
\textbf{Dataset} & \multicolumn{1}{c|}{\textbf{Instruction}} & \multicolumn{1}{c|}{\textbf{Object Tokens}} & \multicolumn{1}{c}{\textbf{Target Object}} \\ \hline\hline
\textbf{R2R} & ``Walk through the kitchen. Go past the sink and stove stand in front of & {[}``kitchen", ``sink", ``stove", ``stand", ``dish", ``table", ``chair"{]} & ``chair"\\
 & the dining table on the bench side." &  \\ \hline
\textbf{SOON} & ``This is a brand new white, rectangular wooden table, which is above a & {[}``book",  ``chair", ``pitcher", ``flower", ``table", ``table"{]} & ``table"\\
 & few chairs, under a pot of flowers. It is in a very neat study with many books." &  \\ \hline
\textbf{REVERIE} & ``Go to the bedroom with the fireplace and bring me the lowest hanging & {[}``bedroom", ``fireplace", ``bed", ``table", ``stand", ``art"{]} & ``art"\\
 & small picture on the right wall across from the bedside table with the lamp on it" & \\
 \bottomrule
\end{tabular}%
\caption{\small \textbf{Object Parsing Examples.} For each dataset, object tokens are extracted from the instructions. Target objects are inferred from the instructions using VQA. Words that have similar meanings are unified into a single object word for categorization.}
\label{tab:object-parsing-examples}\vspace{-0.4cm}
\end{table*}

\begin{figure}[t!]{\centering\includegraphics[width=0.9\linewidth]{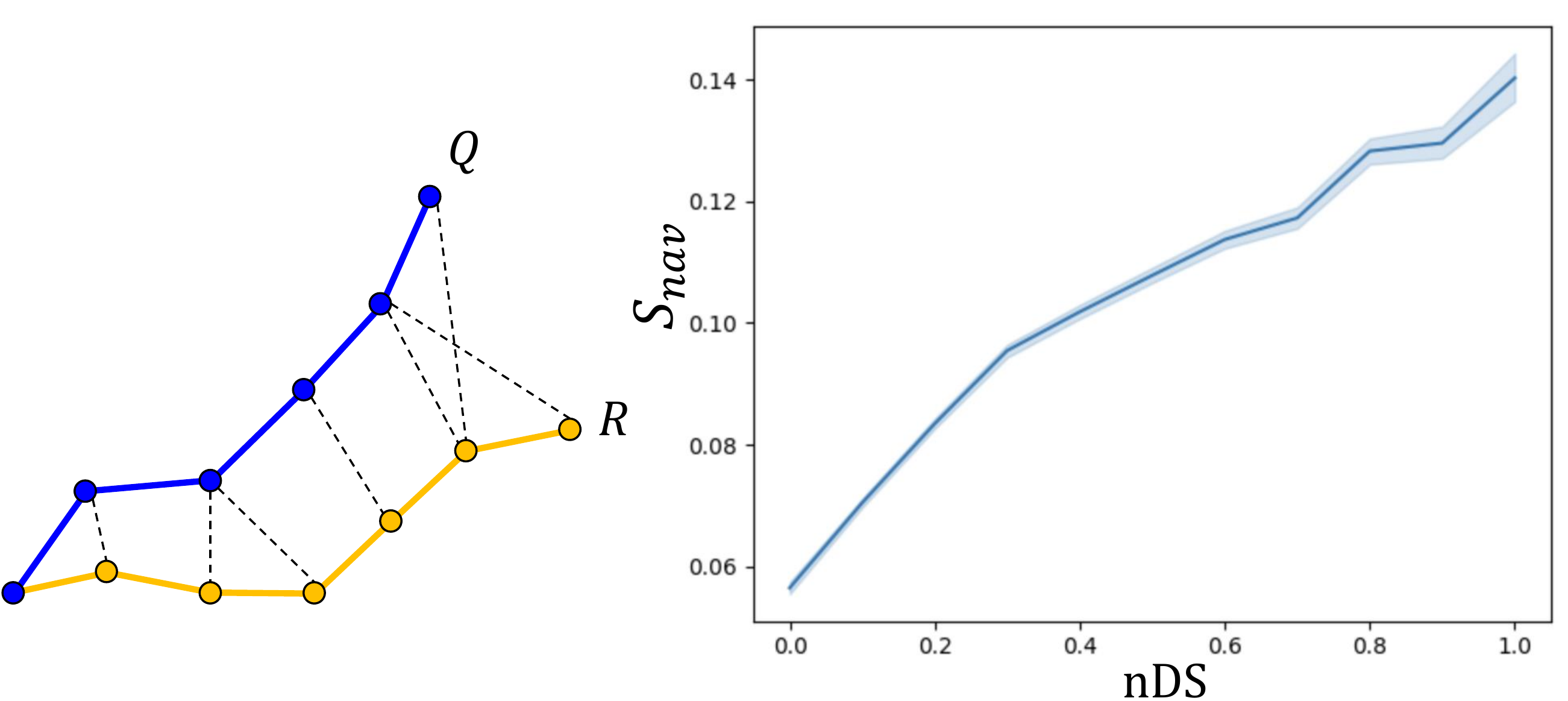}}\centering
\caption{{\textbf{Relationship between navigation score ($S_{nav}$) and normalized distance sum (nDS) in R2R.} We measure navigation scores for augmented trajectories which include both successful and failed trajectories. $R$ and $Q$ illustrate an example case of a ground truth trajectory and a query trajectory. Maximum hop of a query trajectory is 15. Trajectories with high nDS scores also have high navigation scores.
}}\label{fig:navigation-score-plot} 
\end{figure}

\vspace{-0.4cm}
\subsubsection{Navigation Score}
To compare local goal candidates, we design a navigation score of a corrected trajectory $\tau'$ as equation~\ref{eq:navigation-score}. This metric can also be interpreted as a weighted correlation coefficient among SOS features and object tokens weighted by the similarities between them.

{\footnotesize
\begin{align}\label{eq:navigation-score}
    \begin{split}
        S_{nav}(\mathcal{T}') = {{\sum\limits_{i=1}^{B}\sum\limits_{j=1}^{t'} ({\hat\delta(w^o_i)\over{|\hat\delta(w^o_i)|}}\cdot {\vec{S}'_j\over{|\vec{S}'_j|}})((\hat\delta(w^o_i)-\hat\delta(\overbar{w}^o))\cdot(\vec{S}'_j-\overbar{\vec{S}}'))}\over{\sqrt{{{t'}\over B}\cdot\sum\limits_{i=1}^{B} (\hat\delta(w^o_i)-\hat\delta(\overbar{w}^o))^2\sum\limits_{j=1}^{t'} (\vec{S}'_j-\overbar{\vec{S}}')^2}}}
    \end{split}
\end{align}
}

\renewcommand{\baselinestretch}{1.5}
Figure~\ref{fig:navigation-score-plot} shows the relationship between the navigation score and an evaluation metric in the R2R navigation task. Both metrics measure how similar the current trajectory is to the ground truth trajectory. We generate \num[group-separator={,}]{49986} augmented trajectories with an average length of $8.23 m$ based on \num[group-separator={,}]{5596} ground truth trajectories. To generate various samples, we separate each augmented trajectory $(v_1, v_2, ..., v_t)$ into $t$ augmented trajectories $(v_1), (v_1, v_2), ...$, and $(v_1, v_2, ..., v_t)$. The final \num[group-separator={,}]{421383} augmented trajectories include trajectories with 1 to 15 nodes and include both successful and unsuccessful trajectories. We classify the trajectories with the normalized distance sum (nDS) between ground truth trajectory $R$ and a query trajectory $Q$ as follows:
{\scriptsize	
\begin{align}\label{eq:nds}
    \begin{split}
        \text{nDS}(R, Q) = \exp{\Bigg(-\frac{\displaystyle \sum_{v_i \in R}{\min_{u_j \in Q}{d(v_i, u_j)}} + \sum_{u_j \in Q}{\min_{v_i \in R}{d(u_j, v_i)}}}{\frac{|R|+|Q|}{2}d_{success}}\Bigg)},
    \end{split}
\end{align}
}
\renewcommand{\baselinestretch}{1}

\noindent which requires the ground truth information of $R$. $d(u,v)$ denotes the geodesic distance between two nodes, $u$ and $v$, and $d_{success}$ denotes the success distance. The plot in Figure~\ref{fig:navigation-score-plot} shows a linear relationship between the nDS and the navigation score. The results imply that the proposed navigation score effectively scores the augmented trajectories even though it only relies on the given target instruction and observation from the augmented paths, without any location information about the nodes on the ground truth trajectory. 

\subsection{Implementation Details}
We use ViT-B/16 \cite{dosovitskiy2020vit} pretrained on ImageNet to extract
features from the viewpoint panoramic images. We use pretrained LXMERT \cite{tan2019lxmert} for the language encoder and cross-modal transformer. We implement the mode selector as a two-layer feed-forward network.


\section{Experiment Setup}\label{sec:experiment-setup}
\subsection{Dataset Statistics}
\noindent\textbf{R2R.} 
The average length of instructions is 32 words. The average path length of the ground truth trajectory of each instruction is six steps. The number of train, val seen, val unseen, and test episodes are \num[group-separator={,}]{14025}, 1020, 2349, and 4173.

\noindent\textbf{SOON.} \fontdimen2\font=2.0pt
The average path length of the ground truth trajectory of each instruction is four to seven steps. The number of train, validation seen instruction, validation seen house, validation unseen house episodes are 3085, 245, 195, and 205.
\fontdimen2\font=2.5pt

\noindent\textbf{REVERIE.} 
  The average path length of the ground truth trajectory of each instruction is
  9.5 steps. The number of train, val seen, val unseen, and test episodes are \num[group-separator={,}]{10466}, 1423, 3521, and 6292.

\subsection{Data Preprocessing}
To calculate reference SOS features, we preprocess object tokens from language instructions. Using a pretrained visual question answering (VQA) model \cite{antol2015vqa} with the question \textit{"What is the target object? Answer in one word."}, we extract target objects from the instructions in R2R and REVERIE datasets. For SOON dataset, the target object names are already given. After extracting target objects, we perform object parsing for the instructions as shown in Table~\ref{tab:object-parsing-examples}. The final object tokens are sorted by order of appearance in the instructions for R2R and REVERIE. For SOON, considering that the full instruction is divided into 5 parts: object name, object attribute, object relationship, target area, and neighbor areas, we sort the object tokens by reversed order of sentences.

\subsection{Baselines}
\noindent Seq2Seq \cite{anderson2018vision} uses sequence-to-sequence action prediciton to generate actions from the agent trajectory.
Speaker-Follower \cite{fried2018speaker} uses the speaker model to augment natural language instructions and evaluate the candidate action sequence. 
FAST \cite{ke2019tactical} uses both local and global signals to look forward the unobserved environment during exploration and backtrack to the originally visited nodes when needed.
SMNA \cite{ma2019self} uses visual-textual co-grounding module that encodes the past instructions and the instructions and actions to be done. SMNA also uses a progress monitor to estimate the current progress of the agent relative to the total instructions.
Regretful-Agent \cite{ma2019regretful} improves SMNA via two modules. The regret module decides whether to continue to explore or rollback to previous state by a learned policy, and the progress marker decides the direction the agent should head to by selecting visited nodes with progress estimates.
RCM \cite{wang2019reinforced} applies reinforcement learning to enforce the global matching between the agent trajectory and the given natural language instruction. Via cycle-reconstruction reward, RCM allows the agent to comprehend the natural language instruction and penalize paths that do not match with the given instructions.
FAST-MATTN \cite{qi2020reverie} introduces a Navigator-Pointer model to both navigate to the target point and to localize the object from the navigation point according to the language guidance.
AuxRN \cite{zhu2020vision} introduces four auxiliary tasks that help learning the navigation policy: a trajectory retelling task, a progress estimation task, an angle prediction task, and cross-modal matching task, and improves navigation success by aligning representations in these unseen domains with seen domain.
HAMT \cite{chen2021history} uses transformer instead of a recurrent unit to predict actions from a long-range trajectory of observations and actions.
Airbert \cite{guhur2021airbert} uses ViLBert \cite{lu2019vilbert} to measure the correlation between the language instructions and
the viewpoint trajectories.
VLN$\circlearrowright$BERT \cite{hong2021vln} adds a recurrent unit in the transformer to predict the action from the trajectory.
SIA \cite{Lin_2021_CVPRsia} first pretrains the agent to learn the cross-modality between object grounding task and scene grounding task, and then generates real action sequences with memory-based attention.
SSM \cite{Wang_2021_CVPR-structured-scene} integrates information during exploration and constructs a scene memory and chooses the most probable node among visited nodes during backtracking. 
GBE \cite{zhu2021soon} models the navigation state as a graph and explores the environment based on the navigation graph.
DUET \cite{chen2022think} uses two models, a local encoder and a global map planner, to fuse the local observations and coarse scale encoding
for planning actions.\\

\vspace{-0.3cm}
\section{Navigation Experiments}
\label{sec:navigation-experiments}
\vspace{-0.2cm}
In this section, we analyze the evaluation results of navigation experiments with different evaluation metrics. The results are provided in the paper.

\begin{table*}[ht]
\setlength{\tabcolsep}{4.07pt}
\fontsize{7}{8}\selectfont
\begin{tabular}{c|c|c|cccc|cccccc|cccc}
\toprule
\textbf{Methods} & \textbf{Memory} & \textbf{Exploit} &
\multicolumn{4}{c|}{\textbf{Val Seen}} & \multicolumn{6}{c|}{\textbf{Val Unseen}} & \multicolumn{4}{c}{\textbf{Test Unseen}} \\
 &  &  & SR$\uparrow$ & SPL$\uparrow$ & OSR$\uparrow$ & TL$\downarrow$ & SR$\uparrow$ & SPL$\uparrow$ & FSR$\uparrow$ & FSPL$\uparrow$ & OSR$\uparrow$ & TL$\downarrow$ & SR$\uparrow$ & SPL$\uparrow$ & OSR$\uparrow$ & TL$\downarrow$ \\ \hline\hline
Human & - & - & - & - & - & - & - & - & - & - & - & - & 81.51 & 53.66 & 86.83 & 21.18 \\ \hline
Seq2Seq ~\cite{anderson2018vision} & Rec & \redx & 29.59 & 24.01 & 35.70 & 12.88 & 4.20 & 2.84 & 2.16 & 1.63 & 8.07 & 11.07 & 6.88 & 3.99 & 10.89 & \textbf{3.09} \\
VLN$\circlearrowright$BERT ~\cite{hong2021vln} & Rec & \redx & 51.79 & 47.96 & 53.90 & 13.44 & 30.67 & 24.90 & 18.77 & 15.27 & 35.02 & 16.78 & 29.61 & 23.99 & 32.91 & 15.86 \\
RCM \cite{wang2019reinforced} & Rec & \redx & 23.33 & 21.82 & 29.44 & 10.70 & 9.29 & 6.97 & 4.89 & 3.89 & 14.23 & 11.98 & 7.84 & 6.67 & 11.68 & 10.60 \\
\rowcolor{Gray}SMNA \cite{ma2019self} & Rec & \textbf{\bluetext{homing}} & 41.25 & 39.61 & 43.29 & \textbf{7.54} & 8.15 & 6.44 & 4.54 & 3.61 & 11.28 & \textbf{9.07} & 5.80 & 4.53 & 8.39 & 9.23 \\
FAST-MATTN \cite{qi2020reverie} & Rec. & \redx & 50.53 & 45.50 & 55.17 & 16.35 & 14.40 & 7.19 & 7.84 & 4.67 & 28.20 & 45.28 & 19.88 & 11.61 & 30.63 & 39.05 \\
HAMT \cite{chen2021history} & Seq & \redx & 43.29 & 40.19 & 47.65 & 12.79 & 32.95 & 30.20 & 18.92 & 17.28 & 36.84 & 14.08 & 30.40 & 26.67 & 33.41 & 13.62 \\
SIA \cite{Lin_2021_CVPRsia} & Seq. & \redx & 61.91 & 57.08 & 65.85 & 13.61 & 31.53 & 16.28 & 22.41 & 11.56 & 44.67 & 41.53 & 30.80 & 14.85 & 44.56 & 48.61 \\
Airbert \cite{guhur2021airbert} & Seq. & \redx & 47.01 & 42.34 & 48.98 & 15.16 & 27.89 & 21.88 & 18.23 & 14.18 & 34.51 & 18.71 & 30.28 & 23.61 & 34.20 & 17.91 \\
DUET \cite{chen2022think} & Top. Map & \redx & 71.75 & 63.94 & \textbf{73.86} & 13.86 & 46.98 & 33.73 & 32.15 & 23.03 & 51.07 & 22.11 & \textbf{52.51} & 36.06 & \textbf{56.91} & 21.30 \\ \hline
\rowcolor{LightCyan}\textbf{Meta-Explore (Ours)} & Top. Map & \textbf{\bluetext{local goal}} & 71.68 & 63.90 & 73.79 & 13.84 & 47.49 & 34.03 & \textbf{32.32} & 23.30 & \textbf{51.21} & 22.12 & - & - & - & - \\
\rowcolor{LightCyan}\textbf{Meta-Explore$^\ast$ (Ours)} & Top. Map & \textbf{\bluetext{local goal}} & \textbf{71.89} & \textbf{65.71} & 73.44 & 13.03 & \textbf{47.66} & \textbf{40.27} & 32.15 & \textbf{27.21} & 50.55 & 18.48 & 51.18 & \textbf{44.04} & 53.8 & 10.23 \\
\bottomrule
\end{tabular}
\caption{\small \textbf{Comparison and evaluation results of the baselines and our model in REVERIE Navigation Task.}
}\vspace{-0.4cm}
\begin{center}{\footnotesize Gray shaded rows describe hierarchical navigation baselines. Three memory types: Rec\! (recurrent), Seq\! (sequential), and Top. Map\! (topological map)}\end{center}
\label{tab:reverie-baseline_results}
\end{table*}
\subsection{Detailed Analyses in R2R}
\noindent\textbf{Navigation Error (NE).}
Navigation error (NE) is measured as the average distance between the final location of the agent and the target location of episode in meters. Because each episode is recorded as success if NE is less than $3m$, NE is strongly related with the success rate. Meta-Explore shows the lowest NE in the val seen and test unseen splits of the R2R navigation task. The results imply that hierarchical exploration with local goal search helps the agent arrive to the target location closer than other baselines. 

\noindent\textbf{Trajectory Length (TL).}
Among all the R2R navigation baselines, Seq2Seq shows the lowest TL. However, Seq2Seq shows low success rate and low SPL in all data splits. Compared to navigation baselines with SPL higher than $50\%$ in the test split, VLN$\circlearrowright$BERT, SMNA, and HAMT-e2e show lower TL than Meta-Explore. However, all three of these methods show a lower success rate, SPL, and NE than Meta-Explore. According to R2R \cite{anderson2018vision}, train episodes show a wide range of average trajectory length from $5m$ to $25m$, while the test episodes have an average trajectory length of $9.93m$. This implies that the agent is trained with longer trajectories than the test split trajectories, thereby the navigation policy might have learned to navigate longer paths better than shorter paths.

\subsection{Detailed Analyses in SOON}
\noindent\textbf{Oracle Success Rate (OSR).}
In the SOON navigation task, Meta-Explore achieves the highest OSR in the test split while it does not improve the OSR in the val seen instruction and val seen house splits. The proposed method shows a significant generalization result compared to the baselines. AuxRN shows the highest OSR in both the val seen instruction split and the val seen house split as $78.5\%$ and $97.8\%$, respectively, but shows the OSR in the test unseen split as $11.0\%$. On the other hand, Meta-Explore shows OSR as $96.0\%$, $52.7\%$, and $48.7\%$ in the val seen instruction, val seen house, and test unseen splits, respectively. Meta-Explore outperforms AuxRN on OSR by $442.7\%$ in the test split. 

\noindent\textbf{Object Grounding Performance (FSPL).} 
Following \cite{zhu2021soon}, we measure the object grounding performance with the target finding success weighted by path length (FSPL). Although Meta-Explore show the highest success rate and SPL in the val seen instruction and test splits, it does not improve FSPL over baseline methods. We expect to achieve better performance on FSPL if the agent uses the SOS features as deterministic clues to find the target object at the end of each episode. 

\begin{figure*}[t!]{\centering\includegraphics[width=1.0\linewidth]{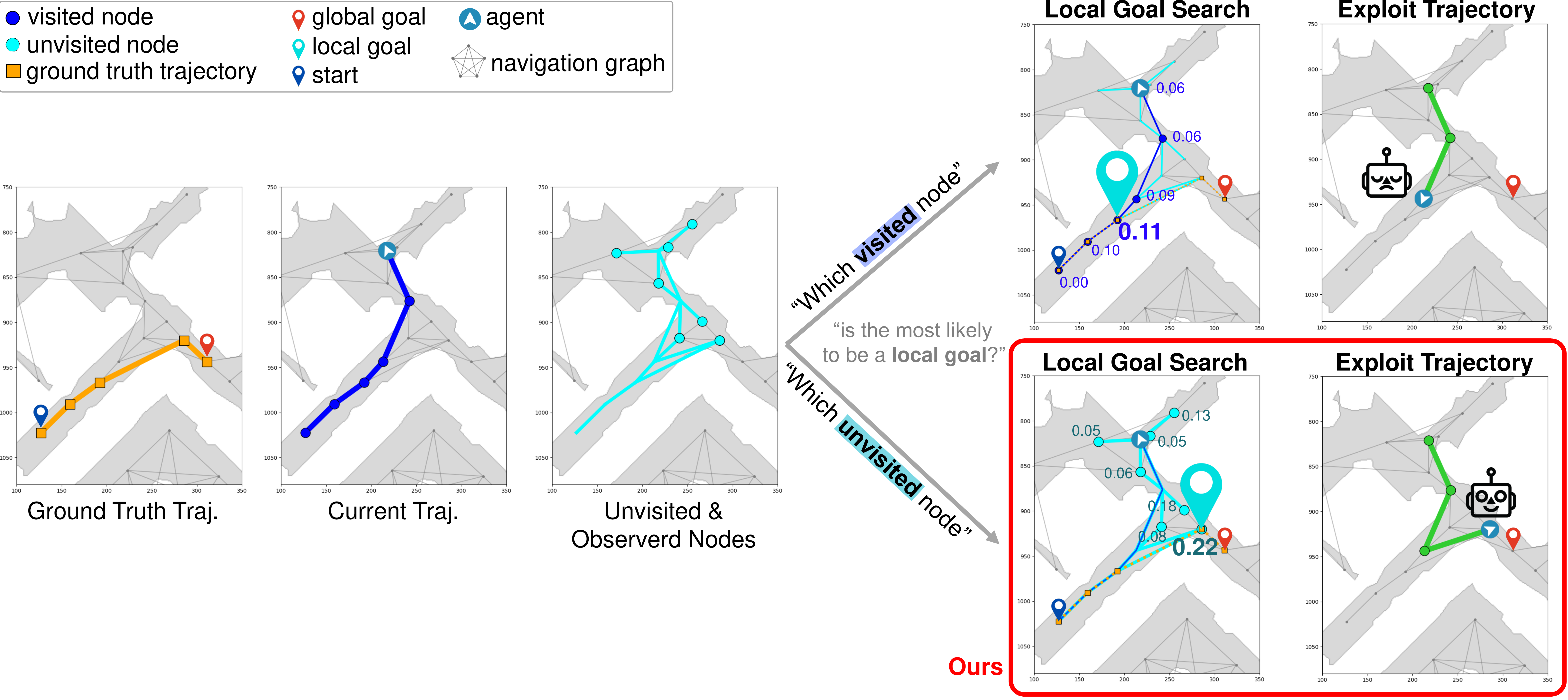}}\centering
\caption{{\textbf{Local goal search scenarios in R2R.} Ground truth trajectory (orange) and current trajectory at time $t=6$ (blue) are shown in the left. Traj. denotes trajectory. The number next to each node denotes the navigation score $S_{nav}$ of the shortest path trajectory from the start node to the corresponding node. If the local goal is chosen from the previously visited nodes, the local goal becomes the node with $S_{nav}=0.11$. If the local goal is chosen from the unvisited but observed nodes, the local goal becomes the node with $S_{nav}=0.22$.
}}\label{fig:local-goal-search-scenario1}
\end{figure*}
\subsection{Evaluation Results in REVERIE benchmark}
Table~\ref{tab:reverie-baseline_results} compares Meta-Explore with the baselines in the REVERIE navigation task. While the proposed method does not improve performance in the val seen split, Meta-Explore outperforms other baselines in the val unseen on success rate, SPL, FSR, FSPL, and OSR. However, the improvement of performance is lower than the improvements shown in R2R and SOON benchmarks. We found 252 meaningless object categories (e.g., verbs, adjectives, and prepositions) and 418 replaceable object categories (e.g., typographical errors and synonyms) in the REVERIE dataset. 10.7\% and 41.2\% of a total of 46,476 words in the bounding box dataset correspond to meaningless and replaceable object categories, respectively. Because our exploitation method utilizes object-based parsing of the given instruction to match with the detected object categories, the effectiveness of the proposed method is lessened due to inaccuracies and inconsistencies in the dataset. We expect to have higher performance if the mistakes in the dataset are fully fixed. 
To provide evidence for this hypothesis, we evaluate Meta-Explore with a modified dataset, which is partially fixed. Typographical errors are fixed and words that have similar meanings are unified into a single object category. For instance, `blackboard', `whiteboard', and `bulletin' are all unified into `board'. The results are shown as the performance of \textbf{Meta-Explore$^\ast$} in Table~\ref{tab:reverie-baseline_results}. The results imply that the proposed method can effectively enhance the SPL by classifying the detected objects correctly, using the modified dataset. 

Comparison between exploitation policies in the REVERIE navigation task is shown in Table~\ref{tab:localgoal_compare_results-reverie}. Among the four exploitation methods: random, spatial, spectral local goal search and homing, spectral-domain local goal search shows the highest performance. The results in Table~\ref{tab:localgoal_compare_results-reverie} are consistent with the results in R2R and SOON.
\begin{table}[htb!]
\renewcommand{\arraystretch}{1.0}
\setlength{\tabcolsep}{2.2pt}
\fontsize{7}{8}\selectfont
\begin{tabular}{c|ccccc|ccccc}
\toprule
\textbf{Local} & \multicolumn{5}{c|}{\textbf{Val Seen}} & \multicolumn{5}{c}{\textbf{Val Unseen}}\\
\textbf{Goal} & SR$\uparrow$ & SPL$\uparrow$ & FSR$\uparrow$ & OSR$\uparrow$ & TL$\downarrow$ & SR$\uparrow$ & SPL$\uparrow$ & FSR$\uparrow$ & OSR$\uparrow$ & TL$\downarrow$ \\ \hline\hline
Oracle  & 79.20 & 64.17 & 62.83 & 84.05 & 18.53 & 59.07 & 38.23 & 40.36 & 66.86 & 26.71 \\ \hline
Random   & 0.21 & 0.04 & 0.00 & 20.31 & 46.02 & 1.11 & 0.18 & 0.34 & 26.70 & 0.05 \\
Homing   & 68.45 & 50.54 & 55.24 & 73.23 & 17.95 & 43.60 & 28.25 & 29.59 & 49.28 & 25.64 \\
Spatial  & 67.53 & 40.21 & 54.25 & 70.91 & 26.92 & 40.90 & 23.25 & 27.61 & 45.84 & 26.92 \\
\rowcolor{LightCyan}Spectral  & \textbf{71.68} & \textbf{63.90} & \textbf{57.34} & \textbf{73.79} & \textbf{13.84} & \textbf{47.49} & \textbf{34.03} & \textbf{32.32} & \textbf{51.21} & \textbf{22.12} \\
\bottomrule
\end{tabular}\vspace{-0.2cm}
\caption{\small Comparison of Exploitation Policies. (REVERIE)
}
\label{tab:localgoal_compare_results-reverie}\vspace{-0.2cm}
\end{table}

\begin{figure}[t!]{\centering\includegraphics[width=1.0\linewidth]{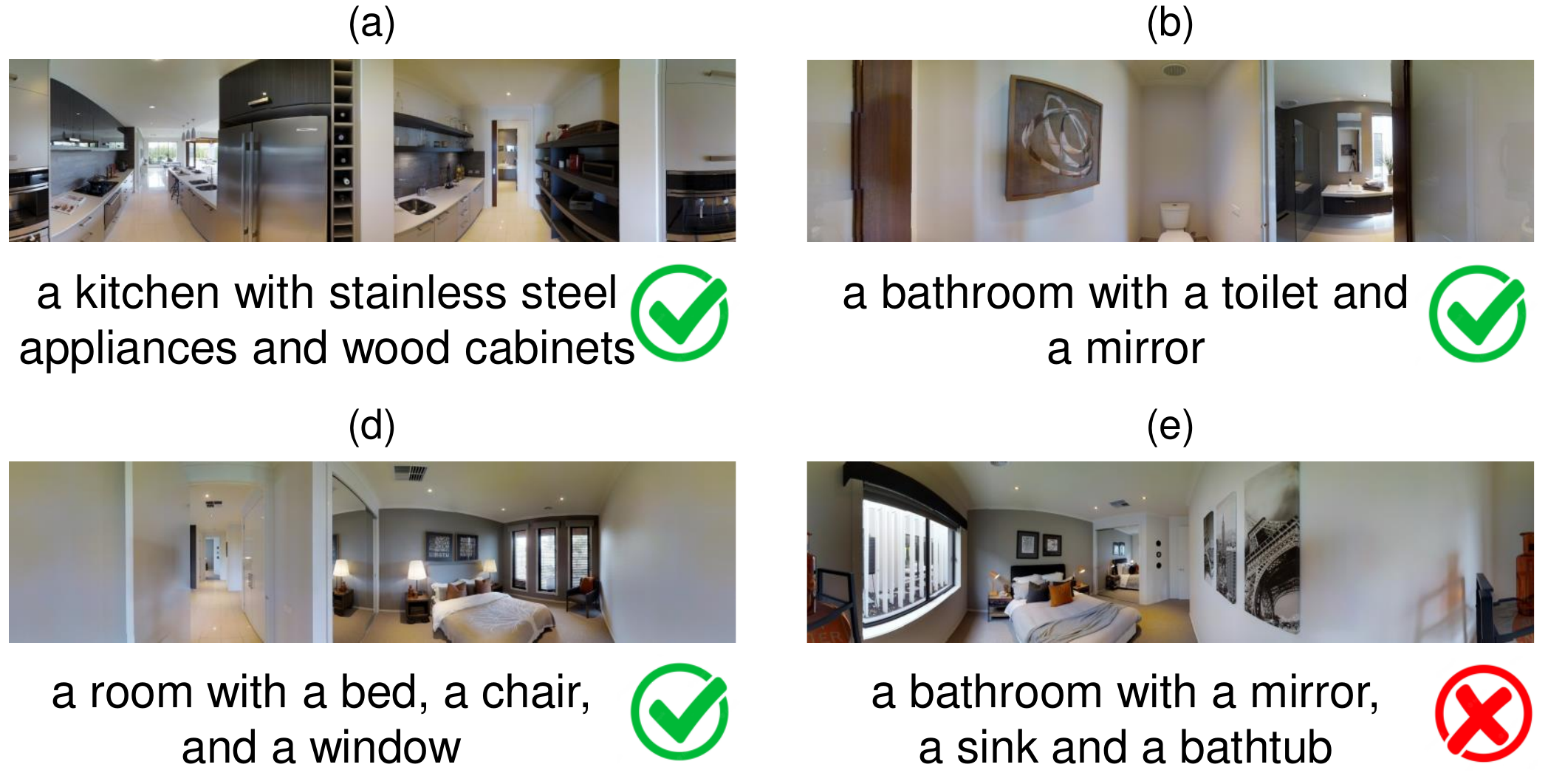}}\centering
\caption{{\textbf{Sample image captions.} (a), (b), (c), and (d) show captions that successfully describe the scenes. (e) and (f) shows failure cases of caption generation. For successful language-triggered hierchical exploration, image captions should correctly describe the scenes. However, current image captioning methods often generates misdescribed captions, thereby leading to a low navigation performance.
}}\label{fig:image-caption-example} \vspace{-0.5cm}
\end{figure}

\subsection{Local Goal Search}
In this section, we provide sample local goal search scenarios. Figure~\ref{fig:local-goal-search-scenario1} shows two scenarios of choosing the local goal when the agent moves to a wrong direction. The agent is given an instruction \textit{``Turn right and turn right again after the desk on the right. Wait next to the cabinets and microwave."}. In both scenarios, we assume that the agent chooses the local goal as the node with the highest navigation score among the possible candidates. If the local goal is chosen from the previously visited nodes, the agent has to move back toward the explored regions. In contrast, if the local goal is chosen from unvisited but observered nodes, the agent can choose a local goal which is close to the global goal. The two scenarios imply that the local goal search in Meta-Explore is more effective than exploitation methods that return the agent to a previously visited node.

\section{Ablation Study}\label{sec:ablation-study}
\subsection{Language-triggered Hierarchical Exploration}

In the proposed method, the target instruction and local goal candidates are compared in spectral-domain using SOS features. Since semantic information can also be expressed in language-domain, we further experiment with the local goal search method using synthesized language captions from visual observations in the R2R navigation task. We compare three types of representation domains: spatial, spectral, and language, which are implemented as panoramic RGB image embeddings, SOS features, and sentence embeddings, respectively. To compare features in different domains, we transfer the source domain to another using augmentation or cross-domain similarity.

\subsubsection{Implementation Details}
We address that the agent can use image captioning to extract contextual information from visual observations such as room type, color, and object placements. To compare local goal candidates and target instruction in language domain, we use pretrained ViT \cite{dosovitskiy2020vit} and GPT-2 \cite{gpt2} to generate the caption for each viewpoint as Figure~\ref{fig:image-caption-example}. The Figure shows four successful cases and two failure cases of image captions. To find a local goal using the generated captions, we calculated the similarities between the captions corresponding to local goal candidates and the target instruction using a fine-tuned sentence transformer `all-MiniLM-L6-v2' \cite{reimers-2019-sentence-bert}. The local goal is chosen as the candidate with the highest similarity. Additionally, we use pretrained CLIP \cite{clip} to evaluate local goal search based on cross-modal similarities between the visual observations of local goal candidates and the target instruction. 

\subsubsection{Experiment Results}
Table~\ref{tab:trigger_compare_results-r2r} shows the evaluation results of the local goal search methods using different target and candidate domains in R2R navigation task. \textit{Nav. Target} denotes the target of VLN, initially given as language. \textit{Lang. Aug.} denotes language captions generated from images. \textit{Spectral Aug.} denotes reference SOS features generated from language instructions. Among the three representation domains, the spectral-domain features enhance navigation performance the most. This implies that hierarchical exploration is most effective when used with spectral visual features. Table~\ref{tab:r2r-baseline_results} and Table~\ref{tab:soon-baseline_results} in the paper also show the improvement of navigation performance by using both hierarchical exploration and spectral visual features over DUET \cite{chen2022thinkshort}, which uses the same ViT-B/16 to extract spatial visual features, resulting in 17.1$\%$ increase in SR and 20.6$\%$ increase in SPL in the SOON test unseen split.

\begin{table}[t!]
\scriptsize
\setlength{\tabcolsep}{8.4pt}
\renewcommand{\arraystretch}{1}
\scalebox{1}{
\begin{tabular}{c|c|cc|cc}
\hline
\multicolumn{2}{c|}{\textbf{Domains}} & \multicolumn{2}{c|}{\textbf{Val Seen}} & \multicolumn{2}{c}{\textbf{Val Unseen}} \\
Nav. Target & Local Goal & SR & SPL & SR & SPL \\ \hline\hline
Lang. & \blackx & 79.92 & 72.79 & 70.63 & 59.81 \\
Lang. & Spatial & 78.84 & 71.96 & 71.05 & 58.86 \\
Lang. & Lang. Aug. & 77.96 & 70.77 & 69.52 & 57.26 \\
\rowcolor{LightCyan}Spectral Aug. & Spectral & \textbf{80.61} & \textbf{75.15} & \textbf{71.78} & \textbf{61.68}\\
\hline
\end{tabular}
}\vspace{-0.2cm}
\caption{\protect\renewcommand{\baselinestretch}{0.88}\small Comparison and evaluation results of the local goal search methods using different target and candidate domains. (R2R)
}
\label{tab:trigger_compare_results-r2r}\vspace{-0.6cm}
\end{table}

\subsection{Image-Goal Navigation in Continuous Domain}\label{image-goal-suppl}
To implicate further applications of Meta-Explore in a continuous domain, we evaluate our method on the photo-realistic Habitat \cite{szot2021habitat} simulator with continuous action space with realistic noises to solve an image-goal navigation task. The objective is to arrive at the target location of the given goal image in an unseen environment. We mainly focus on the effectiveness of hierarchical exploration using local goal search in this experiment. The results are shown in Table~\ref{tab:imagegoal-baseline_results}. 

\begin{figure}[t!]{\centering\includegraphics[width=1.0\linewidth]{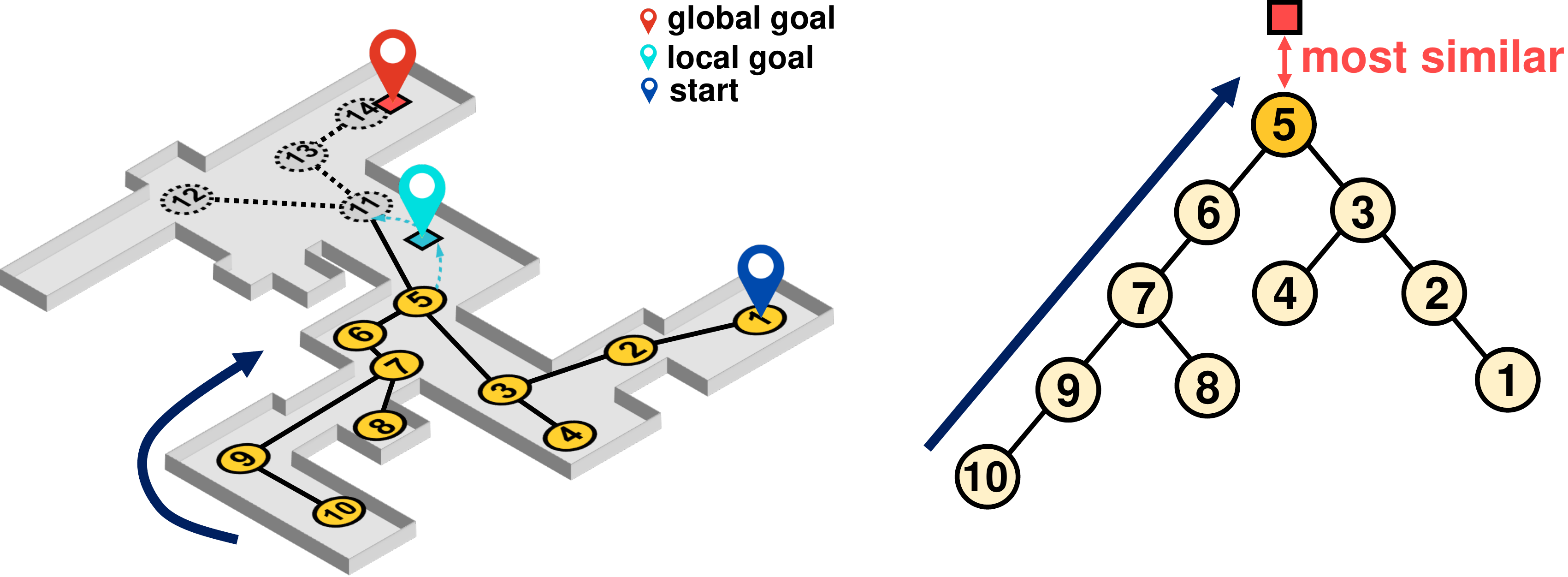}}\centering
\caption{{\textbf{Exploitation by searching a local goal.} In the exploit mode, the agent aims to escape from the stranded local area. It first searches for the most similar node to the goal. Then, it finds an optimal local goal which is unexplored and also similar to the goal image. We use SOS features to compare with the target image.
}}\label{fig:continuous-exploit}
\end{figure}
\subsubsection{Exploration-Exploitation Selection}
We extend Meta-Explore to continuous environments to address the impact of hierarchical exploration in realistic environments. The mode selector decides when to explore and exploit. In the exploration mode, the agent explores around a local area until the meta-controller decides to stop the exploration. The exploration module consists of graph construction module and navigation module. We use recurrent action policy that takes the current and target image features and outputs low-level actions for exploration. We illustrate that the explore-exploit switching decision occurs in stuck scenarios, such as entering a small place or getting stranded in a corner. Figure~\ref{fig:continuous-exploit} shows the overview of exploitation in image-goal navigation by searching a local goal. When the control mode is changed to exploitation mode, the agent returns to the closest previously visited node. Then, the agent finds a local goal among the nodes in the constructed topological map and moves toward the local goal using dijkstra's algorithm \cite{dijkstra1959note}. The local goal is chosen as the node which has the most similar SOS feature with the SOS feature of the target image based on cosine similarity. The agent repeats this explore-exploit behavior until it finds the goal. This explore-exploit switching decision increases the navigation success rate.

\begin{center}
\begin{table*}[t!]
\small{
\renewcommand{\arraystretch}{1.0}
\setlength{\tabcolsep}{6.9pt}
\fontsize{8}{10}\selectfont
\begin{tabular}{c|c|c|cc|cc|cc|cc|cc}
\hline
\textbf{Methods} & \textbf{Exploit} & \textbf{Need} & \multicolumn{2}{c|}{\textbf{Domain}} & \multicolumn{2}{c|}{\textbf{Easy}} & \multicolumn{2}{c|}{\textbf{Medium}} & \multicolumn{2}{c|}{\textbf{Hard}} & \multicolumn{2}{c}{\textbf{Overall}} \\
 &  & \textbf{Pose Info.} & \textbf{spatial} & \textbf{frequency} & \textbf{SR} & \textbf{SPL} & \textbf{SR} & \textbf{SPL} & \textbf{SR} & \textbf{SPL} & \textbf{SR} & \textbf{SPL} \\ \hline
\textbf{VGM \cite{vgm}} & \redx & \textbf{no} & RGBD & \blackx & 0.86 & 0.80 & 0.81 & \textbf{0.68} & 0.61 & 0.46 & 0.76 & \textbf{0.64} \\
\rowcolor{Gray}\textbf{Neural Planner \cite{beeching2020learning}} & \brightgreencheck & global & RGBD & \blackx & 0.72 & 0.41 & 0.65 & 0.39 & 0.42 & 0.27 & 0.60 & 0.36 \\
\rowcolor{Gray}\textbf{NTS \cite{neuralslam}} & \brightgreencheck & global & RGBD & \blackx & 0.87 & 0.65 & 0.58 & 0.38 & 0.43 & 0.26 & 0.63 & 0.43 \\
\rowcolor{Gray}\textbf{ANS \cite{chaplot2020Learning}} & \brightgreencheck & global & RGBD & \blackx & 0.74 & 0.21 & 0.68 & 0.23 & 0.30 & 0.11 & 0.58 & 0.18 \\\hline
\rowcolor{LightCyan}\textbf{Meta-Explore (homing)} & \brightgreencheck & \textbf{local} & RGBD & SOS & 0.82 & 0.61 & 0.83 & 0.61 & 0.70 & \textbf{0.48} & 0.78 & 0.57 \\
\rowcolor{LightCyan}\textbf{Meta-Explore (localgoal)} & \brightgreencheck & \textbf{no} & RGBD & SOS & \textbf{0.94} & \textbf{0.84} & \textbf{0.88} & 0.63 & \textbf{0.71} & 0.18 & \textbf{0.84} & 0.55\\
\hline
\end{tabular}
\caption{\protect\small \textbf{Evaluation results for Image-goal Navigation Task.}\\ (SR: success rate, SPL: success weighted by path length)\\
}
\label{tab:imagegoal-baseline_results}}
\vspace{-0.3cm}
\end{table*}
\end{center}
\vspace{-0.5cm}
\subsubsection{Experiment Details}
We evaluate Meta-Explore in the Gibson dataset \cite{xia2018gibson} with Habitat \cite{szot2021habitat} simulator to solve an image-goal navigation task. Habitat simulator allows the agent to navigate in photo-realistic indoor environments. The exploration policy of the agent is trained using 72 scenes. We evaluate Meta-Explore using 14 unseen scenes. We use panoramic RGBD observations and construct image-based graph memory. To construct a context frequency vector, we detect objects via Mask2Former \cite{cheng2021mask2former} pretrained in ADE-20K dataset \cite{zhou2017scene}, to effectively detect the objects that are generally located in indoor scenes. We use a discrete action space, $\small\{\texttt{stop}, \texttt{move forward}, \texttt{turn left}, \texttt{turn right}\}$ for navigation. With $\small\texttt{move forward}$ action, an agent moves forward by 0.25 m, while $\small\texttt{turn left}$ and $\small\texttt{turn right}$ denotes a ${10}^{\circ}$ rotation, counter-clockwise and clockwise, respectively. The difficulty of each episode is determined by the geodesic distance between the initial and the goal location; \textit{easy}: 1.5 m$\sim$3 m, \textit{medium}: 3 m$\sim$5 m, and \textit{hard}: 5 m$\sim$10 m. The actuation noise model \cite{chaplot2020Learning} is also applied to the agent in order to evaluate in realistic situations. We also demonstrated navigation experiments in the real world using a Jackal robot. The episodes are sampled from simulation point goal episodes with all difficulties; \textit{easy}, \textit{medium} and \textit{hard}. We demonstrate both straight and curved trajectories to evaluate that our model is not task-specific. We used the model only trained in Habitat simulator with Gibson dataset. To collect panoramic RGBD observations, we use one panoramic RGB camera and four front-view RGBD cameras. In order to implement collision avoidance similar to the construction of navigable mesh in Habitat simulator, we implemented a collision avoidance module by clipping the action value based on the depth image observation.
\subsubsection{Baselines}
We compare our image-goal navigation policy with various baselines.
Active Neural SLAM (ANS) constructs a top-down metric map and uses a hierarchical structure consisting of global and local policies. The global policy outputs long-term goals, which are used to generate short-term goals. The local policy uses a geometric path planner to navigate to a short-term goal. NTS \cite{neuralslam} constructs a topological graph during exploration and plans subgoals with graph localization and planning, while navigating to the node with local point goal navigation policy. Neural Planner \cite{beeching2020learning} constructs a graph using an estimated connectivity probability calculated from the neural network.  VGM \cite{vgm} uses unsupervised image-based graph memory representation to compare the similarity between goal image and the current observation image. We adapt VGM for graph construction and local navigation policy. PCL \cite{li2020prototypical} encoder with ResNet18 \cite{he2016deep} backbone network is used as the visual encoder for VGM \cite{vgm}.

\subsubsection{Evaluation Metrics}
We evaluate both success rate ({SR}) and success weighted by inverse path length ({SPL}) \cite{anderson2018evaluation}. An episode is recorded as success if the agent takes a $\texttt{stop}$ action within 1 m of the target location. {SR} is denoted as the number of successes divided by the total number of episodes, $E$. {SPL} is calculated as ${1\over E}\sum_{i=1}^E S_i {l_i\over{\max(p_i, l_i)}}$. $S_i$ denotes the success as a binary value. $p_i$ and $l_i$ denote the shortest path and actual path length for the $i^{th}$ episode. For each task difficulty, {SR} and {SPL} are measured separately.
\subsubsection{Experiment Results}
\begin{figure*}[t!]{\centering\includegraphics[width=0.9\linewidth]{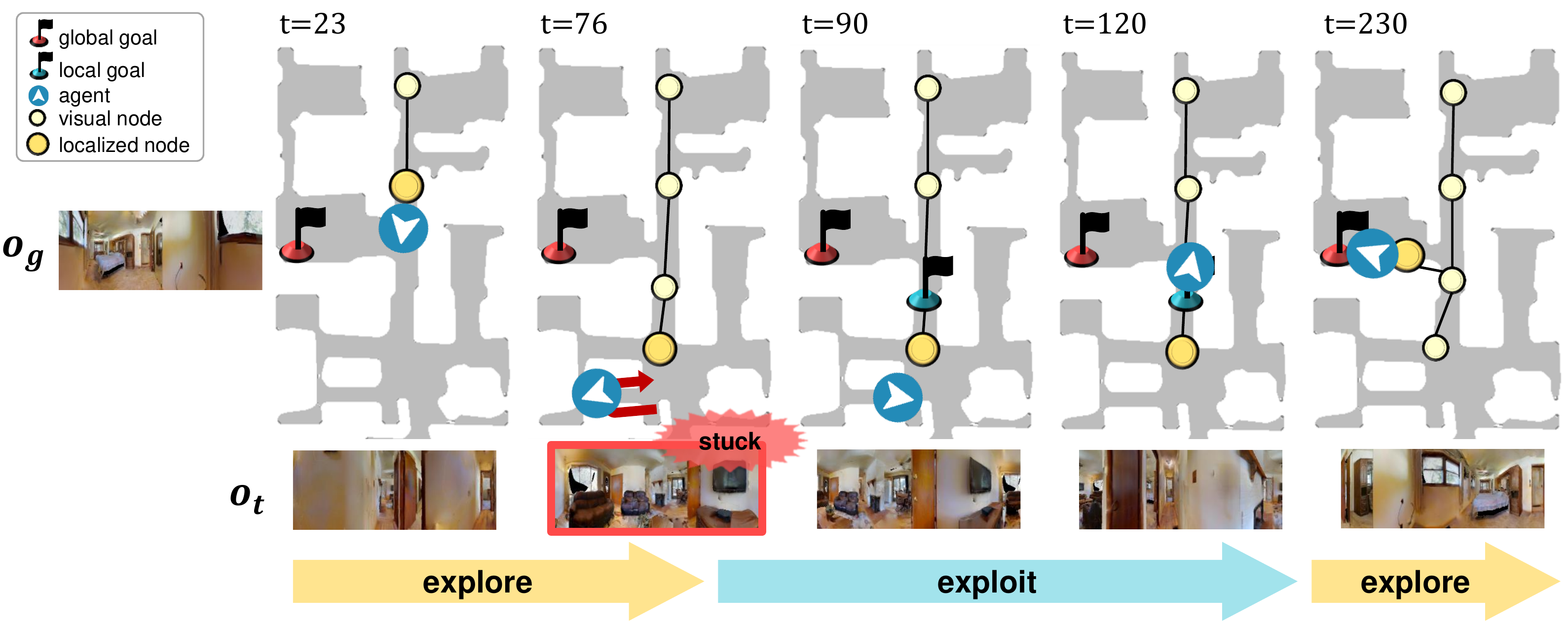}}\centering
\caption{{\textbf{Experiment visualization for image-goal navigation task in continuous environment.} The mode selector detects stuck event at t = 76 and switches the explore mode to exploit mode. Then, the agent returns toward the local goal, which is chosen as a position nearby one of the nodes in the previously constructed graph.
}}\label{fig:continuous-cantwell-scenario}
\end{figure*}

Detailed comparisons with the baseline methods are shown in Table~\ref{tab:imagegoal-baseline_results}. The results show that the continuous version Meta-Explore and SOS features help navigation and the exploitation mode provides corrections for misled exploration or undesirable actions. Compared with the exploration policy baseline VGM \cite{vgm}, Meta-Explore shows an enhancement in the overall success rate by $10.5\%$. The results imply that local goal search helps the agent escape from the current location when the agent recurrently explores a local area but cannot find the target location. Exploitation can reduce unnecessary exploration and help the agent reach the target goal before the maximum time horizon. Among two methods of exploitation, local goal search outperforms homing, presumably because of the noisy actuation model used in the simulator. Due to the noisy actions, the agent can hardly return to a previously visited location by directly reversing the action sequence.

Comparing our method with other graph-based hierarchical navigation methods, Meta-Explore outperforms ANS, Neural Planner, and NTS in the success rate. Our model shows lower performance in SPL for \textit{hard} episodes while the success rate is higher than the baselines. This implies that the exploitation mode of the proposed method allows the agent to explore more uncovered areas. Meanwhile, the proposed method appears to yield a positive impact for \textit{easy} episodes, with the increase on both success rate by $9.3\%$ and SPL by $5.0\%$. Specifically, our method outperforms ANS in terms of both success rate and SPL across all episodes. When compared to Neural Planner and NTS, our approach shows better performance in both success rate and SPL for \textit{easy} and \textit{medium} episodes, while outperforming Neural Planner and NTS in success rate for \textit{hard} episodes.
On the other hand, the proposed method shows lower SPL for \textit{hard} episodes than NTS and Neural Planner. This implies that Meta-Explore tends to explore uncovered areas in both successful and unsuccessful episodes, which could be the result of using the SOS features to understand scenes. Comparing the proposed method using different exploitation methods (homing and local goal search) shows that searching for a local goal leads the agent to better escape from a local area. Figure~\ref{fig:continuous-cantwell-scenario} shows a simple scenario of image-goal navigation using Meta-Explore. The mode selector detects a regretful situation when the agent is recurrently exploring a local area but cannot find the target location. Hierarchical exploration via local goal search helps the agent overcome the situation and move toward the global goal in fixed time.

\subsection{VLN in Continuous Domain}

Image-goal navigation results in complex settings (continuous environments with noisy actions, max$\sim$300 steps) imply that our model can be transferred to long-horizon VLN with noisy actions. We further extend the proposed method in continuous environments to solve the VLN-CE \cite{krantz_vlnce_2020} task. In the VLN-CE \cite{krantz_vlnce_2020} task, our agent constructs a topological map by using Conti-CMA \cite{hong2022bridging} as a baseline to find reachable nodes (i.e., waypoints) and reuses the map in the exploitation mode. 
We compare our continuous version Meta-Explore with various navigation baselines\footnote{$^\dagger$ indicates reproduced results.}: VLN-CE \cite{krantz_vlnce_2020}, HCM \cite{irshad2021hierarchical}, SASRA \cite{Irshad2021sarsa}, and Conti-CMA$^\dagger$ \cite{hong2022bridging}. We evaluate algorithms using the success rate (SR), success weighted by inverse path length (SPL), oracle success rate (OSR), trajectory length (TL), and navigation error (NE), following the definitions of the evaluation metrics in the paper.

\begin{table}[tbp!]
\captionsetup{font=footnotesize}
\resizebox{0.48\textwidth}{!}{%
\begin{tabular}{c|c|c|ccccc}
\toprule
\textbf{Methods} & \textbf{Memory} & {\textbf{Exploit}} & SR$\uparrow$ & SPL$\uparrow$ & OSR$\uparrow$ & TL$\downarrow$ & NE$\downarrow$ \\ \hline\hline
VLN-CE \cite{krantz_vlnce_2020} & Rec & \redx & 32 & 30 & 40 & 8.64 & 7.37 \\
HCM$^\dagger$ \cite{irshad2021hierarchical} & Rec & \redx & - & - & 43 & 15.61 & 8.93 \\
SASRA \cite{Irshad2021sarsa} & Semantic Map & \redx & 24 & 22 & - & \textbf{7.89} & 8.32 \\
Conti-CMA$^\dagger$ \cite{hong2022bridging} & Top. Map & \redx & 41 & 35 & 51 & 10.90 & 6.20 \\
\rowcolor{LightCyan}\textbf{Meta-Explore (Ours)} & Top. Map & local goal & \textbf{49} & \textbf{38} & \textbf{54} & 14.88 & \textbf{4.25} \\
\bottomrule
\end{tabular}
}
\caption{\footnotesize Evaluation results in the VLN-CE val unseen split.
}
\label{tab:vlnce-baseline_results}\vspace{-0.3cm}
\end{table}

\subsubsection{Experiment Results}
Results in Table~\ref{tab:vlnce-baseline_results} show that our method outperforms other baselines by at least $19.5\%$ in the success rate, $8.6\%$ in SPL, and $5.9\%$ in OSR. We excluded the results of HCM for SR and SPL because HCM measures SR, SPL using oracle stop in the official code, which is not allowed in other baselines. We address that our model can be transferred to long-horizon (max. step 300) VLN with noisy actions in complex settings, as demonstrated by image-goal navigation results in Sec.~\ref{image-goal-suppl}.

\end{document}